\documentclass{article}

\usepackage{arxiv}

\usepackage[utf8]{inputenc} 
\usepackage[T1]{fontenc}    
\usepackage{hyperref}       
\usepackage{url}            
\usepackage{booktabs}       
\usepackage{amsfonts}       
\usepackage{nicefrac}       
\usepackage{microtype}      
\usepackage{lipsum}
\usepackage{amsbsy}
\usepackage{amsmath}
\usepackage{amsthm}
\usepackage{comment}
\usepackage{algorithm2e}
\usepackage{array}
\usepackage{enumerate}

\usepackage{caption}
\usepackage{subcaption}
\usepackage{graphicx}
\usepackage{titlesec}

\makeatletter
\newcommand\subsubsubsubsection{\@startsection{paragraph}{4}{\z@}{-2.5ex\@plus -1ex \@minus -.25ex}{1.25ex \@plus .25ex}{\normalfont\normalsize\bfseries}}

\makeatother

\newtheorem{remark}{Remark}

\newtheorem{theorem}{Theorem}
\newtheorem{corollary}{Corollary}

\newtheorem{lemma}{Lemma}

\title{Vector Summaries of Persistence Diagrams for Permutation-based Hypothesis Testing}

\author{
    Umar Islambekov\\
    Department of Mathematics and Statistics\\
    Bowling Green State University\\
    \texttt{iumar@bgsu.edu} \\
        \And
    Hasani Pathirana\\
    Department of Mathematics and Statistics\\
    Bowling Green State University\\
    \texttt{pathirh@bgsu.edu} \\
}

\begin{document} 

\maketitle

\begin{abstract}
    Over the past decade, the techniques of topological data analysis (TDA) have grown into prominence to describe the shape of data. In recent years, there has been increasing interest in developing statistical methods and in particular hypothesis testing procedures for TDA. Under the statistical perspective, persistence diagrams -- the central multi-scale topological descriptors of data provided by TDA -- are viewed as random observations sampled from some population or process. In this context, one of the earliest works on hypothesis testing focuses on the two-group permutation-based approach where the associated loss function is defined in terms of within-group pairwise bottleneck or Wasserstein distances between persistence diagrams \cite{robinson2017hypothesis}. However, in situations where persistence diagrams are large in size and number, the permutation test in question gets computationally more costly to apply. To address this limitation, we instead consider pairwise distances between vectorized functional summaries of persistence diagrams for the loss function. In the present work, we explore the utility of the Betti function in this regard, which is one of the simplest function summaries of persistence diagrams. We introduce an alternative vectorization method for the Betti function based on integration and prove stability results with respect to the Wasserstein distance. Moreover, we propose a new shuffling technique of group labels to increase the power of the test. Through several experimental studies, on both synthetic and real data, we show that the vectorized Betti function leads to competitive results compared to the baseline method involving the Wasserstein distances for the permutation test.   
\end{abstract}



\section{Introduction}
\label{sec:introduction}
Topological data analysis (TDA) provides tools to describe the shape structure underlying data \cite{edelsbrunner2010computational, carlsson2009topology}. One such widely used tool is the theory of \emph{persistent homology} which uses algebraic techniques to quantify the global topology and local geometry of data \cite{edelsbrunner2000topological, zomorodian2005computing, edelsbrunner2008persistent}. A standard topological descriptor provided by persistent homology is a \emph{persistence diagram} (PD) which is commonly computed via a medium of increasing family of combinatorial objects, called simplicial complexes, built on top of data points. 

In applications, PDs, viewed as signatures of the data encoding its global topology and local geometry, are utilized within statistical or machine learning frameworks \cite{chazal2021introduction}. A direct way of doing this is to introduce a notion of distance between PDs, such as the bottleneck or Wasserstein distances \cite{mileyko2011probability, cohen2007stability}, and apply a distance-based method \cite{chung2022persistence}. However, the fact that the space of PDs cannot be turned into a Hilbert space \cite{bubenik2019embeddings, CarBau}, which is the input space for a wide class of machine learning methods, hinders the direct usage of PDs in practice. To address this challenge, one can use PDs indirectly within a learning task by mapping them into some Hilbert space \cite{chung2022persistence}. For example, the mapping of PDs into a Hilbert space can be achieved via a suitable kernel function to be used within a kernel-based machine learning method \cite{chen2015statistical, kusano2016persistence, li2014persistence, reininghaus2015stable,carriere2017sliced}. Alternatively, PDs can be summarized as elements of $\mathbb R^d$ \cite{bubenik2015statistical,adams2017persistence,chung2022persistence,umeda2017time,chan2022computationally,atienza2020stability,richardson2014efficient}. A common way of extracting vector summaries from PDs is first to map them into a space of functions and then vectorize the corresponding functional summaries using a grid of scale values. In recent years, summarizing PDs by means of kernel functions and vectorization has been an active research domain in TDA.  

Integration of TDA-based summaries within machine learning methods has proven very successful in a broad range of applications where data has intrinsic underlying shape structure (see e.g., \cite{hensel2021survey} for a survey of topological methods in machine learning). By contrast, applications of TDA to statistics have not received the same level of attention and remain relatively less explored. A statistical approach to TDA views a given collection of PDs as a random sample drawn from some population or process and addresses problems such as deriving consistency and convergence results for TDA-based summaries, constructing confidence regions and performing hypothesis testing \cite{chazal2021introduction,10.1093/jrsssc/qlad024,berry2020functional}.  

In the present paper, we aim to expand the scope of the permutation test introduced in \cite{robinson2017hypothesis} to test whether two groups of PDs are sampled from the same population. The associated loss function of the permutation test involves computing within-group pairwise bottleneck or Wasserstein distances between PDs. The computation of these distances requires finding optimal bijections between pairs of PDs and has the running time that grows cubically with the size of diagrams. Moreover, the total run-time cost of the test has a quadratic dependence on the number of PDs. Although efficient algorithms exit for approximating the Wasserstein distance \cite{kerber2017geometry,chen2021approximation}, the hypothesis testing procedure in question can become of limited use when PDs are large in size and number. To deal with the issue of high computational cost, we instead consider pairwise distances between vectorized functional summaries of PDs which have significantly lower time complexity (see the simulation experiment in Section \ref{sec:cost} for comparison). Though we focus on and adopt for this work the Betti function\footnote{also known as \emph{Betti curve}}, which is one of the simplest univariate functional summaries extracted from a PD, our approach provides a general framework to incorporate any other vector summary in the hypothesis testing procedure used in \cite{robinson2017hypothesis}. We explore the permutation-based hypothesis testing procedure under the proposed approach in terms of performance and computational gain by conducting a range of experiments involving various types of input data (both synthetic and real) as well as different ways to construct a nested sequence of simplicial complexes on top of the data. Furthermore, we modify the permutation technique of the group labels and empirically demonstrate that the modification consistently leads to the increased power of the test (for more details, see Sections \ref{sec:HTP} and \ref{sec:HT_simulated_data}).

Many of the univariate functional summaries of PDs, including the Betti function, are vectorized by evaluating them at each point of a super-imposed grid. We propose an alternative vectorization scheme, where the Betti function is vectorized by averaging it using integration (see Section \ref{sec:theory} for more details).  

Of the key properties of PDs is their stability against certain class of perturbations in the input data \cite{cohen2007stability,chazal2014persistence,cohen2010lipschitz,mileyko2011probability}. Stability results have also been derived for many of the vector summaries of PDs such as persistence landscapes (PL) \cite{bubenik2015statistical} and persistence images (PI) \cite{adams2017persistence}. We derive stability results for the Betti function and its vector summary with respect to the Wasserstein distance\footnote{The proposed vectorization of the Betti function and the mentioned stability results are also presented in our recent work on predicting anomalies in time-varying graphs (for a preprint of the paper, see \cite{islambekov2023fast}). For completeness, we provide the details of the new vectorization and the proofs of the stability results in Section \ref{sec:theory} with minor modifications.}.   

The rest of the paper is organized as follows. Section \ref{sec:theory} covers the basic theory behind TDA, the new vectorization method for the Betti function and the associated stability results. In Section \ref{sec:HTP}, we review the hypothesis testing procedure for PDs introduced in \cite{robinson2017hypothesis}. In Section \ref{sec:experiments}, we present our experimental findings and comparisons with other vector summaries in terms of performance and run-time cost. Section \ref{sec:conclusion} summarizes the main contributions of the paper and highlights directions for future research.

\section{Theory}
\label{sec:theory}

\subsection{Basics of TDA}
Persistent homology is one of the main tools used in TDA to extract shape information from data \cite{edelsbrunner2000topological, zomorodian2005computing, edelsbrunner2008persistent}. The basic idea behind persistent homology is to build a nested sequence of topological spaces, often in the form of \emph{simplicial complexes} (indexed by a scale parameter), on top of data points and keep a record of the appearance and disappearance of various topological features at different scale resolutions. In this context, these topological features are "holes" of different dimensions -- connected components, loops, voids, and their higher-dimensional versions whose emergence and subsequent disappearance are tracked using a concept of homology from algebraic topology. An increasing sequence of simplicial complexes (called a \emph{filtration}) serves as a medium for gaining insight into the underlying shape structure of the data which is lost during sampling \cite{nanda2013simplicial}.

\begin{figure}[t]
    \centering
    \includegraphics[scale=0.25]{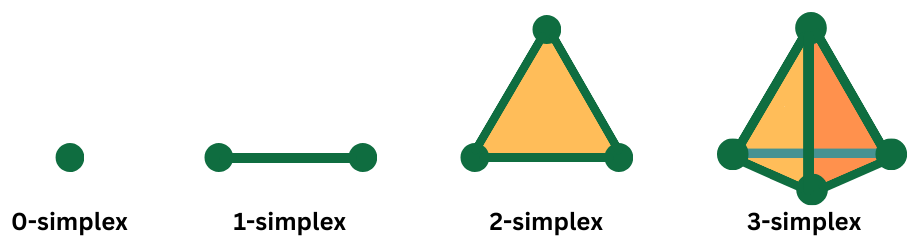}
    \vspace{0.2cm}
    \caption[Geometric representation of simplices]{\small Geometric representation of simplices. Simplices serve as building blocks for simplicial complexes which are used to approximate the underlying shape structure lost during sampling.}
    \label{fig:simplices}
\end{figure}

An (abstract) simplicial complex is a collection $\mathcal K$ of subsets of some finite set $\mathbb{X}$ (called a \emph{vertex set}) such that $\{x\}\in \mathcal K$ for all $x\in \mathbb{X}$ and if $\tau\in \mathcal K$ then $\sigma\in \mathcal K$ for all $\sigma\subseteq\tau$. If $\tau \in \mathcal K$ consists of $k+1$ points, it is called a $k$-simplex (or a simplex of dimension $k$). The dimension of a simplicial complex is the maximum dimension of its simplices. It is convenient to identify a 0-simplex with a vertex, a 1-simplex with an edge and a 2-simplex with a triangle and so on (see Figure \ref{fig:simplices}). 

Building a filtration is an important step in obtaining topological information from data. A general way to construct a filtration is by using sublevelsets of a monotone function defined over a simplicial complex \cite{nanda2013simplicial}. Let $\mathcal{K}$ be a simplicial complex and $g:\mathcal K\rightarrow\mathbb{R}$ a monotone function in the sense that if $\sigma,\tau\in \mathcal{K}$ with $\sigma\subseteq\tau$, then $g(\sigma)\leq g(\tau)$. Let $\{c_1,c_2,\ldots,c_N\}$ be the increasing sequence of distinct values of $g$. The sublevelset filtration of $\mathcal K$ induced by $g$ is a nested family of simplicial complexes:
$$
\emptyset \subseteq S_1(g) \subseteq  S_2(g) \subseteq \ldots \subseteq S_N(g)=\mathcal K, 
$$
where $S_i(g)=\{\sigma\in \mathcal K\ |\ g(\sigma)\leq c_i\}$ and $N$ is called the length of the filtration. 

In applications, the choice of $g$ to construct filtered simplicial complexes depends on the nature of the data set (such as size and dimension), computational considerations and what kinds of topological/geometric features of the data one would like to highlight \cite{nanda2013simplicial}. For example, if $\mathbb X$ is a finite set of points lying in some metric space $(M,\rho)$ and $\mathcal K_{\mathbb{X},d}$ is the simplicial complex of dimension $d$ containing all subsets of $\mathbb X$ of size at most $d+1$, then the sublevelset filtration of $\mathcal K_{\mathbb{X},d}$ induced by $g(\sigma)=\max_{x,y\in \sigma}\rho(x,y)$ is called the \emph{Vietoris-Rips} filtration. The length of this filtration is equal to the number of distinct pairwise distances among the points of $\mathbb X$. Note that the number of simplices in $\mathcal K_{\mathbb{X},d}$ increases exponentially with the size of $\mathbb X$. Therefore, to reduce the computational cost, in practise one commonly constructs a shorter filtration before reaching $\mathcal K_{\mathbb{X},d}$ by stopping at the scale value reasonably less than the maximum distance \cite{nanda2013simplicial}. The \emph{\v{C}ech} filtration is obtained if  $g(\sigma)=\min\{r\geq 0\ |\ \exists x\in M \hbox{ such that } \rho(x,y)\leq r \hbox{ for all } y\in\sigma\}$. While the \v{C}ech filtration has more desirable theoretical properties, in applications the Vietoris-Rip filtration is often preferred due to its ease of construction and fast implementation. 

If $\mathbb X$ is a finite abstract set and $g(\sigma)=\max_{x\in\sigma}f(x)$, where $f:\mathbb X\rightarrow\mathbb R$, the corresponding filtration is called the \emph{lower-star} filtration. Thus, to construct the lower-star filtration, it suffices to have any real-valued function $f$ defined over the vertex set $\mathbb X$ which does not have to lie in a metric space. 
When $\mathbb X\subseteq \mathbb R^d$ and $g(\sigma)=\min\{r\geq 0\ |\ \cap_{x\in\sigma} B_x(\sqrt{r})\cap V_x\neq \emptyset \}$, where $B_x(\sqrt{r})$ is the closed ball centered at $x$ with radius $\sqrt{r}$ and $V_x = \{z \in \mathbb{R}^d\ |\ \|z-x\|_2 \leq \|z-y\|_2 \hbox{ for all } y\in \mathbb X\}$ (Voronoi cell of $x$), one obtains filtered \emph{alpha} complexes. Observe that unlike the Vietoris-Rip and \v{C}ech complexes, the dimension of an alpha complex can never go beyond $d$. Moreover, the so-called \emph{nerve theorem} is applicable to filtered alpha complexes. However, these benefits come at the expense of increased computational cost which makes alpha complexes less practical in higher dimensions (i.e., when $d$ is large) \cite{nanda2013simplicial}. In section \ref{sec:experiments}, we illustrate the usage for most of these filtrations. 

A standard summary of various topological features that appear and disappear as we traverse a filtration is called a \emph{persistence diagram} (PD). It is a multiset of points in $\mathbb{R}^2$ where each point $(b,d)$ represents a topological feature that is "born" at scale value $b$ and "dies" at scale value $d$. Note that a PD is a multiset since it may contain multiple features  with the same birth and death values. The scale values vary over the range of the function $g$ which induces the filtration. A $k$-dimensional PD consists of points corresponding to topological features of homological dimension $k$ (0 if connected components, 1 if one-dimensional holes (or loops), 2 if two-dimensional holes (or voids), etc). If the homological dimension of a PD is understood or irrelevant in a given context, we will omit it to reduce notation. The features with low persistence values (i.e. $d-b$) tend to correspond to noise, while those having high persistence are more likely to reflect true topological features of the underlying shape. Thus, a PD can be viewed as a topological fingerprint of the data it is computed from \cite{chazal2021introduction}. 

Shape structures underlying two sets of data points can be compared by introducing a distance metric between the computed PDs. The $L_p$ $q$-Wasserstein distance \cite{kerber2017geometry}, $p,q\geq 1$, between two finite persistence diagrams $D$ and $\tilde D$ is defined as
\begin{equation}
  \label{w-distance}
d_{pq}(D,\tilde D)=\Big(\inf_{\gamma:D\rightarrow \tilde D} \sum_{u\in D} \|u-\gamma(u)\|_{p}^q \Big)^{1/q},  
\end{equation}

where $\|\cdot\|_p$ is the $L_p$ norm on $\mathbb{R}^2$ and $\gamma$ is a bijection with $D$ and $\tilde{D}$ being enlarged to additionally contain points of the diagonal line $\Delta=\{(b,d)\in \mathbb{R}^2\ |\  b=d\}$ with infinite multiplicity. Thus, $\gamma$ pairs an off-diagonal point either with another off-diagonal point from the other diagram or with its own orthogonal projection on the diagonal line $\Delta$. If $p=\infty$, the metric is simply called the $q$-Wasserstein distance. If both $p$ and $q$ are infinite, one obtains the \emph{bottleneck} distance (where the summation is replaced by the supremum). 

\subsection{Summary functions of persistence diagrams}
Since the Wasserstein distance involves finding the optimal bijection, the overall computational cost can get very high if PDs are large in size or in number. One way of handling this limitation is to extract functional summaries from PDs, discretize them to obtain vectors in $\mathbb{R}^n$ and use the $L_p$ distance between the vectors instead of the Wasserstein distance. Vectorization methods such as \emph{persistence landscapes} \cite{bubenik2015statistical}, \emph{persistence images} \cite{adams2017persistence} and \emph{Betti functions} \cite{chazal2021introduction,chung2022persistence} fall under this approach. Below we briefly review their constructions and vectorization methods.

The $k$th order landscape function of a persistence diagram $D=\{(b_i,d_i)\}_{i=1}^N$ is defined as 
$$\lambda_k(t) = k\hbox{max}_{1\leq i \leq N} \Lambda_i(t), \quad k\in\mathbb{N},
$$
where $k\hbox{max}$ returns the $k$th largest value and 
\begin{equation*}
    \Lambda_i(t) = \left\{
        \begin{array}{ll}
            t-b_i & t\in [b_i,\frac{b_i+d_i}{2}] \\
            d_i-t & t\in (\frac{b_i+d_i}{2},d_i]\\
            0 & \hbox{otherwise}
        \end{array}
    \right.
\end{equation*}
To obtain a finite-dimensional vector, the landscape function is evaluated at each point of an increasing sequence of scale values $\{t_1,t_2,\ldots,t_n\}$: 
\begin{equation} \label{vec:PL}
(\lambda_k(t_1),\lambda_k(t_2),\ldots,\lambda_k(t_n))\in\mathbb{R}^n.  
\end{equation}

A persistence image is a vector summary of the \emph{persistence surface} function 
$$ \rho(x,y)=\sum_{u\in T(D)}f(u)\phi_u(x,y),$$
where $T(D)=\{(b_i,p_i)\}_{i=1}^N$ with $p_i=d_i-b_i$ (persistence), $\phi_u$ is a differentiable probability distribution (usually a Gaussian) with mean $u=(u_x,u_y)$ and $f$ is a suitable weighting function with zero values along the diagonal $\Delta$. The persistence surface is vectorized by averaging it over each cell of a superimposed two-dimensional grid:
$$
\Big(\frac{1}{\Delta x_1\Delta y_1}\int_{x_1}^{x_2}\int_{y_1}^{y_2}\rho(x,y)dydx,\ldots,\frac{1}{\Delta x_{n-1}\Delta y_{m-1}}\int_{x_{n-1}}^{x_n}\int_{y_{m-1}}^{y_m}\rho(x,y)dydx\Big)\in\mathbb{R}^{(n-1)(m-1)},
$$
where $\{x_1,x_2,\ldots,x_n\}$ and $\{y_1,y_2,\ldots,y_m\}$ are two increasing sequences of birth and persistence values; $\Delta x_k=x_{k+1}-x_k$ and $\Delta y_j=y_{j+1}-y_j$. 

The Betti function is one of the simplest summary functions and defined as a weighted sum of indicator functions determined by the points of a PD: 
\begin{equation}\label{def:betti}
\beta(t)=\sum_{i=1}^N w(b_i,d_i)I_{[b_i,d_i)}(t),    
\end{equation}

where $I_{[b_i,d_i)}(t)=1$ if $t\in [b_i,d_i)$ and 0 otherwise and $w$ is a weight function. If $w\equiv 1$ and the homological dimension of a PD is $k$, $\beta(t)$ is equal to the count of $k$-dimensional topological features what are born before or at $t$ and still persistent. 

Similar to the persistence landscape function, a common way to vectorize the Betti function is to evaluate it at each point of an increasing sequence of scale values $\{t_1,t_2,\ldots,t_n\}$ and form a vector in $\mathbb{R}^n$. Since the Betti function is easy to integrate, in the present work, we propose to vectorize by averaging it over intervals formed by pairs of successive scale points:
\begin{equation}\label{VAB}
 \Big(\frac{1}{\Delta t_1}\int_{t_1}^{t_2}\beta(t)dt,\frac{1}{\Delta t_2}\int_{t_2}^{t_3}\beta(t)dt,\ldots,\frac{1}{\Delta t_{n-1}}\int_{t_{n-1}}^{t_n}\beta(t)dt\Big)\in\mathbb{R}^{n-1}.   
\end{equation} 

We call this new vector representation a \emph{vector of averaged Bettis} (VAB). Unlike the common approach, the proposed vectorization method considers the behavior of the Betti function over the entire intervals determined by pairs of neighboring scale values. Note that if a grid of scale values is dense enough, the two vector summaries will be very similar but high-dimensional. However, the new vectorization can also be used to compute informative low-dimensional vectors based on a sparse grid of scale values\footnote{See the experiment of Section 3.1 in \cite{islambekov2023fast} which compares the two vectorization methods of the Betti function when the extracted vector summaries are low-dimensional.}.

Among the metrics for comparing PDs that do not involve vectorization is the \emph{sliced Wasserstein} distance \cite{carriere2017sliced}. It utilizes the fact that the Wasserstein distance is fast to compute if the points of PDs all lie on the same line. Thus, for the sliced Wasserstein distance the points of PDs are first projected onto a line through the origin and then the Wasserstein distance is computed between the two sets of projected points. Finally, the Wasserstein distances are averaged over all possible lines via integration.

\subsection{Stability results for the Betti function and vector of averaged Bettis (VAB)}
\label{Stability results for a Betti function and a vector of averaged Bettis (VAB}

PDs are known to be stable to fluctuations with respect to the Wasserstein distance \cite{cohen2007stability,mileyko2011probability}. Likewise, stability is a desirable property of any vector summary of PDs. For example, vector summaries such as persistence landscapes and persistence images have some type of stability guaranties \cite{bubenik2015statistical,adams2017persistence}. In the following, we derive stability results (Theorem \ref{thm:stability} and Corollary \ref{cor:stability}) for the Betti function and its new proposed vectorization with respect to the $L_1$ and $L_2$ 1-Wasserstein distances. We assume that all PDs have finitely many off-diagonal points with finite death values. Under additional assumptions for the weight function $w$ in (\ref{def:betti}), the stability result can be extended to PDs with infinite death values (see Remark \ref{rmk:extention}). First, we prove a technical lemma needed for our main result (Theorem \ref{thm:stability}). 

\begin{lemma}\label{lemma}
Let $w:\mathbb{R}^2\rightarrow \mathbb{R}$ be a differentiable function with $\|w\|_\infty= \sup_{z\in \mathbb{R}^2}|w(z)|<\infty$ and $\|\nabla w\|_\infty =\sup_{z\in \mathbb{R}^2}\|\nabla w(z)\|_2<\infty$. Then for any $u=(b_u,d_u)$, $v=(b_v,d_v)$ with $b_u\leq d_u<\infty$ and $b_v\leq d_v<\infty$, we have 
\begin{equation}\label{lemma:inequality}
\int_{\mathbb{R}}|w(u)I_{[b_u,d_u)}(t)-w(v)I_{[b_{v},d_{v})}(t)|dt \leq \|w\|_\infty\|u-v\|_1+L\|\nabla w\|_\infty\|u-v\|_2,
\end{equation}
where $L$ is a non-negative constant such that $L\geq \max\{d_u-b_u,d_{v}-b_{v}\}$. 

\begin{proof}
The proof consists of three cases. Let $f(t)=|w(u)I_{[b_u,d_u)}(t)-w(v)I_{[b_{v},d_{v})}|$ and note that by the fundamental theorem of calculus for line integrals, $|w(u)- w(v)| \leq \|\nabla w\|_\infty\|u-v\|_2$.
 
 Case 1: $[b_u,d_u)\cap [b_{v},d_{v})= \emptyset$. Without the loss of generality, assume $b_u \leq d_u \leq b_{v}\leq d_{v}$. Then,
 \vspace{-0.1cm}
 \begin{align*}
        \int_{\mathbb{R}}f(t)dt & = \int_{b_u}^{d_u} f(t)dt+\int_{b_{v}}^{d_{v}}f(t)dt\\
        & =\int_{b_u}^{d_u} |w(u)|dt+\int_{b_{v}}^{d_{v}}|w(v)|dt\\
        & \leq \|w\|_\infty |d_u-b_u|+\|w\|_\infty |d_{v}-b_{v}|\\
        & \leq \|w\|_\infty |b_{v}-b_u|+\|w\|_\infty |d_{v}-d_u|\\
        & = \|w\|_\infty (|b_u-b_{v}|+|d_u-d_{v}| )\\
        & = \|w\|_\infty\|u-v\|_1\\
        & \leq \|w\|_\infty\|u-v\|_1+L\|\nabla w\|_\infty\|u-v\|_2.
\end{align*}       

 Case 2:  $[b_u,d_u)\cap [b_{v},d_{v})\neq \emptyset$, $[b_u,d_u)\not\subseteq [b_{v},d_{v})$ and $[b_{v},d_{v})\not\subseteq [b_u,d_u)$. Without the loss of generality, assume $b_u \leq b_{v} \leq d_u \leq d_{v}$. Then, 
 \vspace{-0.1cm}
\begin{align*}
        \int_{\mathbb{R}}f(t)dt & = \int_{b_u}^{b_{v}} f(t)dt+ \int_{b_{v}}^{d_u} f(t)dt +\int_{d_u}^{d_{v}}f(t)dt\\
        & =  \int_{b_u}^{b_{v}} |w(u)|dt+ \int_{b_{v}}^{d_u} |w(u) - w(v)|dt +\int_{d_u}^{d_{v}}|w(v)|dt\\
        & \leq \|w\|_\infty |b_{v}-b_u|+\|\nabla w\|_\infty \|u-v\|_2 |d_u-b_{v}| +\|w\|_\infty |d_{v}-d_u|\\
        & = \|w\|_\infty(|b_u-b_{v}|+ |d_u-d_{v}|) + L \|\nabla w\|_\infty \|u-v\|_2 \\
        & = \|w\|_\infty \|u-v\|_1 +  L\|\nabla w\|_\infty \|u-v\|_2. 
\end{align*}
 
Case 3:  $[b_u,d_u) \subseteq [b_{v},d_{v})$ or $[b_{v},d_{v}) \subseteq [b_u,d_u)$. Without the loss of generality, assume $b_u \leq b_{v} \leq  d_{v} \leq d_u$. Then,
 \vspace{-0.1cm}
 \begin{align*}
        \int_{\mathbb{R}}f(t)dt & = \int_{b_u}^{b_{v}} f(t)dt+ \int_{b_{v}}^{d_{v}} f(t)dt +\int_{d_{v}}^{d_u}f(t)dt\\
        & =  \int_{b_u}^{b_{v}} |w(u)|dt+\int_{b_{v}}^{d_{v}} |w(u) - w(v)|dt +\int_{d_{v}}^{d_u}|w(v|dt\\
        & \leq \|w\|_\infty |b_{v}-b_u|+\|\nabla w\|_\infty \|u-v\|_2 |d_{v}-b_{v}| +\|w\|_\infty |d_u-d_{v}|\\
        & = \|w\|_\infty(|b_u-b_{v}|+ |d_u-d_{v}|) + L \|\nabla w\|_\infty \|u-v\|_2 \\
        & = \|w\|_\infty \|u-v\|_1 +  L \|\nabla w\|_\infty \|u-v\|_2.
\end{align*}

By combining the three cases we finish the proof of the lemma.
\end{proof}
\end{lemma}
\vspace{-0.1cm}
We now state and prove our main stability result. In the context of computing the Wasserstein distance, PDs are additionally supplied with points of the diagonal line $\Delta=\{(b,d)\in \mathbb{R}^2\ |\ b=d\}$. Note that the inclusion of such points does not change the corresponding Betti function.  

\begin{theorem} \label{thm:stability}
Let $D$ and $\tilde{D}$ be two persistence diagrams with finitely many off-diagonal points and finite death values. Let $\beta(t)=\sum_{u\in D}w(u)I_{[b_u,d_u)}(t)$ and ${\tilde{\beta}}(t)=\sum_{v\in \tilde D}w(v)I_{[b_v,d_v)}(t)=\sum_{u\in D}w(\gamma(u))I_{[b_{\gamma(u)},d_{\gamma(u)})}(t)$ be the associated Betti functions, where the weight function $w$ is as in Lemma \ref{lemma} and  $\gamma$ is a bijection between $D$ and $\tilde{D}$. Then,
\begin{equation}\label{thm:main_inequality}
\|\beta-{\tilde{\beta}}\|_{L_1} \leq \|w\|_\infty d_{11}(D,\tilde{D})+L\|\nabla w\|_\infty d_{21}(D,\tilde{D}),    
\end{equation}
\vspace{0.1cm}
where $L=\max_{u\in D_1\cup D_2}(d_u-b_u)$. In particular, if $w\equiv1$ then $\|w\|_\infty=1$, $\|\nabla w\|_\infty=0$ and therefore, we have 
\begin{equation} \label{thm:corrolary_inequality}
\|\beta-{\tilde{\beta}}\|_{L_1} \leq d_{11}(D,\tilde{D}).    
\end{equation}

\begin{proof}
\begin{align*}
  \|\beta-{\tilde{\beta}}\|_{L_1}
  &=\int_{\mathbb{R}}\Big|\sum_{u \in D}w(u) I_{[b_u,d_u)}(t)-\sum_{u \in D}w(\gamma(u)) I_{[b_{\gamma(u)},d_{\gamma(u)})}(t)\Big|dt\\ 
  & \leq \sum_{u \in D}\int_{\mathbb{R}} |w(u)I_{[b_u,d_u)}(t)-w(\gamma(u))I_{[b_{\gamma(u)},d_{\gamma(u)})}(t)|dt \\
  & \leq \|w\|_\infty\sum_{u \in D}  \|u-\gamma(u)\|_1+L\|\nabla w\|_\infty\sum_{u \in D}\|u-\gamma(u)\|_2 \quad \hbox{(by Lemma 2.1)}\\
  & \leq \|w\|_\infty d_{11}(D,\tilde{D})+L\|\nabla w\|_\infty d_{21}(D,\tilde{D}).
\end{align*}
\end{proof}
\end{theorem}

\begin{remark}
If the points of all PDs being considered belong to $\mathbb{R}_{\Delta,m,M}^2=\{(x,y)\ |\ m\leq x\leq y\leq M\}$ for some constants $m,M \in (0,\infty)$, then any differentiable function $w$ defined on the compact set $\mathbb{R}_{\Delta,m,M}^2$ can be taken as a weight function for Theorem \ref{thm:stability}.  
\end{remark}

\begin{remark}\label{rmk:2}
Since $\|u-v\|_2\leq \|u-v\|_1$, $\|u-v\|_1\leq 2\|u-v\|_\infty$ and $\|u-v\|_2\leq \sqrt{2}\|u-v\|_\infty$, the upper bound in (\ref{thm:main_inequality}) can be further bounded above by $ (\|w\|_\infty +L\|\nabla w\|_\infty)d_{11}(D,\tilde{D})$ and $ (2\|w\|_\infty +\sqrt{2}L\|\nabla w\|_\infty)d_{\infty 1}(D,\tilde{D})$.
\end{remark}

\begin{remark}\label{rmk:extention}
In general, a PD may contain points with infinite death values. Note that the associated Betti function is still well-defined. In practice, one may ignore points with infinite death values or replace such values with some global constant. If we choose neither of these options and further assume that $w\equiv C$ (constant) and $\|u-\phi(u)\|_1=|b_1-b_2|$ where $u=(b_1,\infty)\in D$ and $\phi(u)=(b_2,\infty)\in \tilde{D}$ (a common convention adopted by the software tools for TDA), then the upper bound in (\ref{lemma:inequality}) will be $C \|u-v\|_1$ and hence $\|\beta-{\tilde{\beta}}\|_{L_1} \leq C \cdot d_{11}(D,\tilde{D})$. If $D$ and $\tilde{D}$ have unequal number of points with infinite death values, then one of the points is necessarily matched with a point on the diagonal line $\Delta$ which implies $\|\beta-{\tilde{\beta}}\|_{L_1}=d_{11}(D,\tilde{D})=\infty$. If the points with infinite death values are equal in number across $D$ and $\tilde{D}$, then they must be matched among one another by an optimal bijection $\gamma$ and the two distances will both be finite.  
\end{remark}

As proven in \cite{chung2022persistence}, the lack of a stability result for the Betti curve $\beta(t)=\sum_{(b,d)\in D}I_{b<d}(b,d)I_{[b,d)}(t)$ (see Section A.1.1 of the cited paper for more details) does not contradict the claim of Theorem \ref{thm:stability}. In \cite{chung2022persistence}, the Betti curve is considered a member of a broader class of summary functions called \emph{persistence curves}, where the weight functions may additionally depend on $t$, and are assumed to be zero on the diagonal line $\Delta$ and continuous with respect to $t$. The weight function $w(b,d)=I_{b<d}(b,d)$ of the Betti curve is not differentiable on $\mathbb R^2$ and therefore is not suitable for Theorem \ref{thm:stability}. In contrast, we take $w(b,d)\equiv 1$ which proves crucial for proving inequality (\ref{thm:corrolary_inequality}). Note that the two weights differ only on the diagonal $\Delta$ and since both are multiplied by $I_{[b,d)}(t)$, the resulting Betti summaries are the same. 

For some weight functions, the main stability result of \cite{chung2022persistence} (see Theorem 1) and Theorem \ref{thm:stability} provide similar bounds. For example, if $w(b,d)=d-b$ and the points of PDs belong to $\mathbb{R}_{\Delta,m,M}^2$, then according to Theorem 1 in \cite{chung2022persistence}, 
\begin{equation*} \label{ineq:life}
\|\beta-{\tilde{\beta}}\|_{L_1} \leq 4 (M-m) d_{\infty 1}(D,\tilde{D}).    
\end{equation*}
For Theorem \ref{thm:stability}, $\|w\|_\infty\leq M-m$ (assuming $w$ is restricted to $\mathbb{R}_{\Delta,m,M}^2$), $L\leq M-m$ and $\|\nabla w\|_\infty=\sqrt{2}$ implying by Remark \ref{rmk:2}
\begin{align*}
   \|\beta-{\tilde{\beta}}\|_{L_1}&\leq (2+2)(M-m)d_{\infty 1}(D,\tilde{D})\\
  & = 4(M-m)d_{\infty 1}(D,\tilde{D}). 
\end{align*}
Next is a stability result for the vector of averaged Bettis (VAB) which readily follows from Theorem \ref{thm:stability}. 
\begin{corollary}\label{cor:stability}
Under the assumptions of Theorem \ref{thm:stability}, let $\boldsymbol{\beta}$ and $\boldsymbol{\tilde\beta}$ be the corresponding VABs computed using an increasing sequence of equally spaced scale values $t_1,t_2,\ldots,t_n$ via (\ref{VAB}). Then,
$$\|\boldsymbol{\beta}-\boldsymbol{\tilde\beta}\|_{1} \leq \frac{1}{\Delta t}\Big[\|w\|_\infty d_{11}(D,\tilde{D})+L\|\nabla w\|_\infty d_{21}(D,\tilde{D})\Big],
$$
where $\Delta t=t_{i+1}-t_i=\hbox{const}$, $i=1,2,\ldots,n-1$. In particular, if $w\equiv1$ then $\|w\|_\infty=1$, $\|\nabla w\|_\infty=0$ and therefore we have 
$$
\|\boldsymbol{\beta}-\boldsymbol{\tilde\beta}\|_{1} \leq \frac{1}{\Delta t} d_{11}(D,\Tilde{D}).
$$

\begin{proof}
\begin{align*}
\|\boldsymbol{\beta}-\boldsymbol{\tilde\beta}\|_{1} 
&= \sum_{i=1}^{n-1}\Big|\frac{1}{\Delta t_i}\int_{t_i}^{t_{i+1}}\beta(t)dt - \frac{1}{\Delta t_i}\int_{t_i}^{t_{i+1}}\tilde{\beta}(t)dt \Big|\\
& \leq \sum_{i=1}^{n-1}\frac{1}{\Delta t} \int_{t_i}^{t_{i+1}}|\beta(t) -\tilde{\beta}(t)|dt\\
& \leq \frac{1}{\Delta t}\int_{\mathbb{R}}|\beta(t) -\tilde{\beta}(t)|dt\\
& \leq \frac{1}{\Delta t}\Big[\|w\|_\infty d_{11}(D,\tilde{D})+L\|\nabla w\|_\infty d_{21}(D,\tilde{D})\Big] \quad \hbox{(by Theorem \ref{thm:stability})}
\end{align*}
\end{proof}

\end{corollary}

\section{Hypothesis testing procedure for persistence diagrams}
\label{sec:HTP}
In this section we review the hypothesis testing procedure presented in \cite{robinson2017hypothesis}. Let  $\{D_i\}_{i=1}^{n_1}$ and $\{\tilde D_i\}_{i=1}^{n_2}$ be two groups (or sets) of independent PDs. The PDs of each group might be generated from some model or computed from data (e.g., point clouds, images, graphs etc.) via a suitable filtration type. The goal is to test whether the processes behind generating the two groups of PDs are the same (the null hypothesis $H_0$) or different (the alternative hypothesis $H_1$). For example, if PDs of each group arise from point clouds sampled from a 2D shape, the null hypothesis will state that the underlying two shapes are the same.  

Since PDs are not known to be characterized by parametric distributions, the authors of \cite{robinson2017hypothesis} resort to a permutation-based testing procedure which provides an empirical estimate of the null distribution of the test statistic. The paper adopts the following joint loss function involving within-group pairwise distances between PDs as a test statistic:     

\begin{equation} \label{Fpq}
    	F_{pq}(\{D_i\},\{\tilde{D_i}\})=\frac{2}{n_1(n_1-1)}\sum_{i=1}^{n_1}\sum_{j=1}^{n_1}d_{pp}(D_{i},D_{j})^q+
		\frac{2}{n_2(n_2-1)}\sum_{i=1}^{n_2}\sum_{j=1}^{n_2}d_{pp}(\tilde D_{i},\tilde D_{j})^q
		,
\end{equation}
where $d_{pp}$ is the $L_p$ $p$-Wasserstein distance as in (\ref{w-distance}) and $q\geq 1$.  

The permutation-based $p$-value for a given labelling $L_0=\{D_1,\ldots,D_{n_1},\tilde D_1,\ldots,\tilde D_{n_2}\}$ under the loss function $F$ is the proportion of all labelings (permutations of the labels) $L$ for which $F(L)\leq F(L_0)$. The number of all possible labelings is $\binom{n_1+n_2}{n_1}$ which can easily get very large. In this case, one samples uniformly from the set of all permutations which provides an unbiased estimator of the $p$-value. The algorithm for computing the permutation $p$-value is given in Algorithm \ref{alg:p-value} \cite{robinson2017hypothesis}. 

\RestyleAlgo{ruled}
\begin{algorithm}
\caption{Unbiased estimation of the $p$-value \cite{robinson2017hypothesis}}\label{alg:p-value}
\KwData{$n_1+n_2$ PDs with labels $L_0$ in disjoint sets of size $n_1$ and $n_2$, number of permutations $N$, a joint loss function $F$}
\KwResult{Estimate of the $p$-value}
 Initialize $Z$ at $0$\;
 Compute $F(L_0)$ for the observed labels\;
 
 \For{$i=1$ to $N-1$}{
  Randomly shuffle the group labels into disjoint sets of size $n_1$ and $n_2$ to obtain labelling $L$\;
  Compute $F(L)$\;
  \If{$F(L) \leq F(L_0)$}{
   $Z=Z+1$\;
   }{
  }
 }
 $Z=\frac{Z+1}{N+1}$\; 
Output $Z$ /* p-value */
\end{algorithm}

In practice, once the computation of the loss function $F$ is implemented, to run Algorithm \ref{alg:p-value} it suffices to provide the $(n_1+n_2) \times (n_1+n_2)$ matrix of all pairwise distances (which is computed only once), the labelling $L_0$ and the number of permutations $N$.  
   
In Section \ref{sec:experiments}, for Algorithm \ref{alg:p-value} we use the Wasserstein distance $d_{11}$ as in (\ref{Fpq}) or replace it either with the sliced Wasserstein distance or the $L_1$ distance between the corresponding vector summaries extracted from PDs. For example, if $\{\boldsymbol{\beta}_i\}$ and $\{\tilde{\boldsymbol{\beta}}_i\}$ are the vectors of averaged Bettis (VAB) computed from persistence diagrams $\{D_i\}$ and $\{\tilde D_i\}$ respectively using (\ref{VAB}), then $F_{pq}$ taken the form 
\begin{equation} \label{FpqVAB}
    	F_{pq}(\{\boldsymbol{\beta}_i\},\{\tilde{\boldsymbol{\beta}}_j\})=\frac{2}{n_1(n_1-1)}\sum_{i=1}^{n_1}\sum_{j=1}^{n_1}\|\boldsymbol{\beta}_i-\boldsymbol{\beta}_j\|_p^q+
		\frac{2}{n_2(n_2-1)}\sum_{i=1}^{n_2}\sum_{j=1}^{n_2}\|\tilde{\boldsymbol{\beta}}_i-\tilde{ \boldsymbol{\beta}}_j\|_p^q,
\end{equation}
where $\|\cdot\|_p$ is the $L_p$ norm on $\mathbb{R}^2$. 

Moreover, in addition to the standard permutation method of the group labels, we introduce a new shuffling technique with the aim of increasing the power of the test. Under this approach, when permuting the group labels we consider only those permutations which lead to stronger mixing of the labels from different groups. In other words, we maximize the sum of proportions of pairwise distances between PDs of different groups which potentially leads to larger values of the loss function $F$ and hence a smaller p-value. To make the idea more precise, consider permutations where $k$ diagrams from group 1 are put into group 2 and vice versa. Then the sum of proportions of distances between PDs from different groups is 
$$
\frac{k(n_1-k)}{n_1(n_1-1)}+\frac{k(n_2-k)}{n_2(n_2-1)}.
$$
The above quantity (as a function of $k$) is maximized at
\begin{equation} \label{eqn:kmax}
  k_{\hbox{\tiny max}}=\frac{n_1n_2(n_1+n_2-2)}{2(n_1(n_1-1)+n_2(n_2-1))}.
\end{equation}

In practice, following the strong mixing procedure, we take $k=\min(n_1,n_2,k_{\hbox{\tiny max}})$ and consider a subset of the permutations where exactly $k$ PDs are exchanged between the two groups. Note that when the null hypothesis is true, regardless of the permutation method (the standard or strong mixing), the PDs being permuted are drawn from the same underlying population/process.

When the group sizes are equal (i.e., $n_1=n_2$) and $p=2$, the loss function $F$ in (\ref{FpqVAB}) has an interesting connection to the \emph{energy} distance \cite{szekely2005hierarchical} defined by
\begin{equation*}
    E_\alpha=\frac{n_1n_2}{n_1+n_2}[2M_{12}-M_{11}-M_{22}],
\end{equation*}
where $\alpha\in(0,2]$ and
$$
M_{12}=\frac{1}{n_1n_2}\sum_{i=1}^{n_1}\sum_{j=1}^{n_2}\|\boldsymbol{\beta}_i-\tilde{\boldsymbol{\beta}}_j\|_2^\alpha, \quad M_{11}=\frac{1}{n_1^2}\sum_{i=1}^{n_1}\sum_{j=1}^{n_1}\|\boldsymbol{\beta}_i-\boldsymbol{\beta}_j\|_2^\alpha,$$
$$
M_{22}=\frac{1}{n_2^2}\sum_{i=1}^{n_2}\sum_{j=1}^{n_2}\|\tilde{\boldsymbol{\beta}}_i-\tilde{\boldsymbol{\beta}}_j\|_2^\alpha.
$$
Observe that $E_\alpha$ contains not only pairwise within-group distances, but also between-group distances and it can be shown that $E_\alpha\geq0$ \cite{szekely2005hierarchical}. Assuming $n=n_1=n_2$, $p=2$ and $q=\alpha\in[1,2)$, and letting $T=n^2(2M_{12}+M_{11}+M_{22})$ (the sum of all possible pairwise distances), we get
\begin{align*}
    E_\alpha &=\frac{n^2}{2n}\Big[2M_{12}-M_{11}-M_{22}\Big]\\
    &= \frac{n}{2}\Big[\frac{1}{n^2}T-2(M_{11}+M_{22})\Big]\\
    &= \frac{n}{2}\Big[\frac{1}{n^2}T-\frac{4(n-1)}{n}F_{2\alpha}\Big]\\
    &= \frac{T}{2n}-2(n-1)F_{2\alpha}
\end{align*}
Hence $E_\alpha$ is a linear transformation of $F_{2\alpha}$ and since $T$ remains constant under any permutation, Algorithm \ref{alg:p-value} will provide the same p-value if $F_{2\alpha}$ is replaced by $E_\alpha$.

\section{Experiments}
\label{sec:experiments}
The experiments section consists of three main subsections. The first subsection examines the run-time costs of five different methods for computing the matrix of pairwise distances to be used in Algorithm \ref{alg:p-value}. The first one is the baseline method, where PDs are compared via the $L_1$ 1-Wasserstein distance (W). The next three methods involve the $L_1$ distances between respective vector summaries of the vector of average Bettis (VAB), persistence images (PI) and persistence landscapes (PL) computed from the given PDs. The fifth method compares PDs using the sliced Wasserstein distance (SW). The loss function (\ref{Fpq}) is appropriately modified for the methods other than the baseline (e.g., (\ref{FpqVAB}) is used instead of (\ref{Fpq}) for VAB). 

The second subsection explores the power of the hypothesis test (implemented in Algorithm \ref{alg:p-value} of Section \ref{sec:HTP}) for the above five methods (W, VAB, PI, PL, SW) using various simulated data (with and without added noise) and filtration types. The last subsection investigates the performance of the testing procedure on real data. 

In all experiments, the weight function $w$ in (\ref{def:betti}) for the vector of averaged Bettis is set to one. For persistence landscapes we consider only the first landscape function (i.e. $k=1$ in (\ref{vec:PL})). Persistence images are computed using the Gaussian density with standard deviation $\sigma=0.5\times (\hbox{maximum persistence})/(\hbox{grid size})$. The sliced Wasserstein distance is computed (approximately) by taking 10 uniformly spaced directions. All TDA-related computations are performed using the {\tt R} language \cite{R} in conjunction with packages {\tt TDA} \cite{fasy2014introduction,TDA}, {\tt TDAstats} \cite{wadhwa2018tdastats}, {\tt kernelTDA} \cite{kernelTDApackage},  {\tt TDAvec} \cite{islambekov2022package} and {\tt igraph} \cite{igraph}. 

For Algorithm \ref{alg:p-value}, the loss function $F_{pq}$ is computed with $p=q=1$, and both the standard (random) and strong mixing procedures are applied in shuffling the group labels. The null hypothesis $H_0$ (of equality of shapes/processes between the two groups) is rejected if the $p$-value is less than the cutoff $\alpha = 0.05$ (significance level). We consider PDs of different homological dimensions separately and report the results for the dimension which leads to the best outcome.  

\subsection{Performance comparison with respect to computational cost}
\label{sec:cost}

This section focuses on the run-time costs of the five methods (W, VAB, PI, PL, SW) for computing the matrix of pairwise distances. For all simulations, a persistence diagram $D=\{(b_i,d_i)\}_{i=1}^N$ is generated from the process where $b_i\sim \hbox{Unif}(0,1)$ and $d_i=b_i+z_i$ with $z_i \sim \hbox{Beta}(\alpha,\beta)$ for some prespecified $\alpha$ and $\beta$. For the vector summaries VAB, PL and PI the run-time cost additionally includes the time taken to compute these summaries from the given PDs.

In the first simulation study, we generate 50 PDs of size $N$ ranging (per simulation) over $\{100,200,\cdots, 1000\}$ to explore the run-time cost with respect to changes in the size of PDs. In the second study, we vary the number of PDs $n$ from 10 to 100 (with increments of 10) and fix the size of PDs to be 500. Table \ref{table:computational time N} and Table \ref{table: Computational time n} provide the median run-time costs (measured in seconds and computed based on ten repeated simulations) for these two studies. The results show that VAB has the smallest (overall) run-time cost and SW has much smaller cost than the baseline Wasserstein metric which is the costliest to compute. Moreover, we observe that for the vector summaries (i.e. VAB, PI and PL) the computational cost increases approximately linearly in $N$ and $n$. 

\begin{table}[t]
\centering
    \begin{minipage}[c]{.45\textwidth}
      \centering
      \begin{tabular}{cccccc}
		\hline
	    N &	W &	VAB & PL & PI & SW \\   
		\hline
100 & 6.723 & \textbf{0.007} & 0.016 & 0.013 & 0.293 \\ 
  200 & 17.92 & \textbf{0.011} & 0.038 & 0.026 & 0.606 \\ 
  300 & 34.93 & \textbf{0.015} & 0.059 & 0.039 & 1.000 \\ 
  400 & 57.58 & \textbf{0.020} & 0.068 & 0.051 & 1.361 \\ 
  500 & 82.77 & \textbf{0.025} & 0.085 & 0.064 & 1.744 \\ 
  600 & 111.8 & \textbf{0.029} & 0.103 & 0.077 & 2.112 \\ 
  700 & 147.4 & \textbf{0.035} & 0.122 & 0.091 & 2.515 \\ 
  800 & 174.4 & \textbf{0.046} & 0.127 & 0.106 & 2.916 \\ 
  900 & 218.5 & \textbf{0.044} & 0.146 & 0.121 & 3.337 \\ 
  1000 & 245.0 & \textbf{0.050} & 0.179 & 0.143 & 3.744 \\
       \hline
	\end{tabular}
 \vspace{0.1cm}
 \captionof{table}{\small Median run-time costs (measured in seconds) of computing the distance matrix for the five methods (W,VAB,PL,PI,SW) using 50 PDs of size $N$ ranging over \{100,200,\ldots,1000\}.}
	\label{table:computational time N}
    \end{minipage}
   \quad \quad   
    \begin{minipage}[c]{.45\textwidth}
      \centering
     \begin{tabular}{cccccc}	 
        \hline
	     n &	W &	VAB & PL & PI & SW \\ 
		\hline
10 & 2.971 & \textbf{0.005} & 0.013 & 0.013 & 0.061 \\ 
  20 & 11.82 & \textbf{0.010} & 0.027 & 0.025 & 0.274 \\ 
  30 & 29.09 & \textbf{0.015} & 0.044 & 0.039 & 0.607 \\ 
  40 & 50.41 & \textbf{0.019} & 0.062 & 0.051 & 1.084 \\ 
  50 & 87.71 & \textbf{0.026} & 0.086 & 0.069 & 1.807 \\ 
  60 & 117.1 & \textbf{0.037} & 0.104 & 0.075 & 2.425 \\ 
  70 & 159.4 & \textbf{0.034} & 0.117 & 0.086 & 3.329 \\ 
  80 & 207.9 & \textbf{0.046} & 0.137 & 0.120 & 4.279 \\ 
  90 & 260.5 & \textbf{0.046} & 0.153 & 0.116 & 5.504 \\ 
  100 & 323.3 & \textbf{0.056} & 0.174 & 0.123 & 6.796 \\ 
       \hline
	\end{tabular}
  \vspace{0.1cm}
 \captionof{table}{\small Median run-time costs (measured in seconds) of computing the distance matrix for the five methods (W,VAB,PL,PI,SW) using $n$ PDs of size 500, where $n$ ranges over \{10,20,\ldots,100\}.}
	\label{table: Computational time n}
    \end{minipage}
\end{table}
 
\subsection{Hypothesis testing on simulated data }
\label{sec:HT_simulated_data}

This section provides four experimental results regarding the power of the hypothesis test based on different types of simulated data. The first two involve computing PDs from 2D and 3D point clouds (with underlying shape structure) via the Vietoris-Rips and alpha filtrations respectively. The third dataset consists of PDs generated from a process similar to that of Section \ref{sec:cost}. In the fourth dataset, PDs are computed from random graphs via the lower-star filtration.  

All four experiments consider 5 sets of 20 PDs where the first 10 belong to group 1 and the remaining 10 are part of group 2: $\{\mathcal{S}_i\}_{i=1}^5=\{\{D_{ij}\}_{j=1}^{10},\{\tilde{D}_{ij}\}_{j=1}^{10}\}_{i=1}^5$. When $i=1$, the data generating processes behind PDs of the two groups are the same (i.e. the null hypothesis $H_0$ is true). As $i$ gets bigger, the alternative hypothesis $H_1$ is true and the process behind producing PDs of group 2 increasingly starts to differ from that of group 1 which remains fixed. Hence, as $i$ gradually increases, the two groups of PDs, in principle, become easier to distinguish and therefore, we expect the power of the hypothesis test to increase as well (at a higher or lower rate depending on the absence or presence of noise in the data). 

In terms of vectorization of summary functions of PDs, we use a (one-dimensional) grid of size 100 for the Betti functions and persistence landscapes, and a $20 \times 20$ grid for the persistence images. For Algorithm \ref{alg:p-value}, under the standard permutation method of the group labels, we randomly select 10000 out of $\binom{20}{10}=184756$ possible permutations. When the strong mixing procedure (introduced in Section \ref{sec:HTP}) is applied, since the group sizes $n_1=n_2=10$, by (\ref{eqn:kmax}) $k_{\hbox{\tiny max}}=5$ PDs from each group are mixed with each other which leads to $\binom{10}{5}\cdot\binom{10}{5}=64504$ permutations of which we again randomly select 10000. For each value of $i$, the power of the test is computed based on 1000 independent repeated simulations. When $i=1$, the null hypothesis $H_0$ is true and therefore the power of the test is equal to the type-I error rate. For $i>1$, the power of the test is equal to 1 - the type-II error rate since the alternative hypothesis $H_1$ is true in this case. 

Finally, we redo all the experiments where the four datasets are generated with added noise. The results will be presented as bar charts where the heights of colored bars correspond to the powers of the hypothesis test for each of the five methods (W, VAB, PI, PL, SW) under the standard permutation method. The additional bars (with transparent background) on top of the colored bars reflect the increase in power under the strong mixing procedure for permuting the group labels. 

\subsubsection{Distinguishing circle and ellipse}
\label{subsec:circle_vs_ellipse}

In this experiment, PDs of group 1, $\{D_{ij}\}_{i=1,j=1}^{5,10}$, are computed (via a Vietoris-Rips filtration) from points clouds of size 50 sampled uniformly from a unit circle. For each $i\in\{1,\ldots,5\}$, PDs $\{\tilde D_{ij}\}_{j=1}^{10}$ of group 2 are computed based on 50 points sampled uniformly from an ellipse with equation $x^2+y^2/(1-r)^2=1$ where $r=0.02(i-1)$. When $r=0$ (or $i=1$), the two underlying shapes are the same and as $r$ increases the corresponding ellipses gradually start to depart (in terms of shape) from the unit circle. For the case with noise, we add Gaussian noise with standard deviation $\sigma=0.05$ to both the $x$ and $y$ coordinates of the points in a point cloud. 

\begin{figure}[t]
\centering
\begin{subfigure}[b]{.4\textwidth}
  \centering 
  \includegraphics[width=\textwidth]{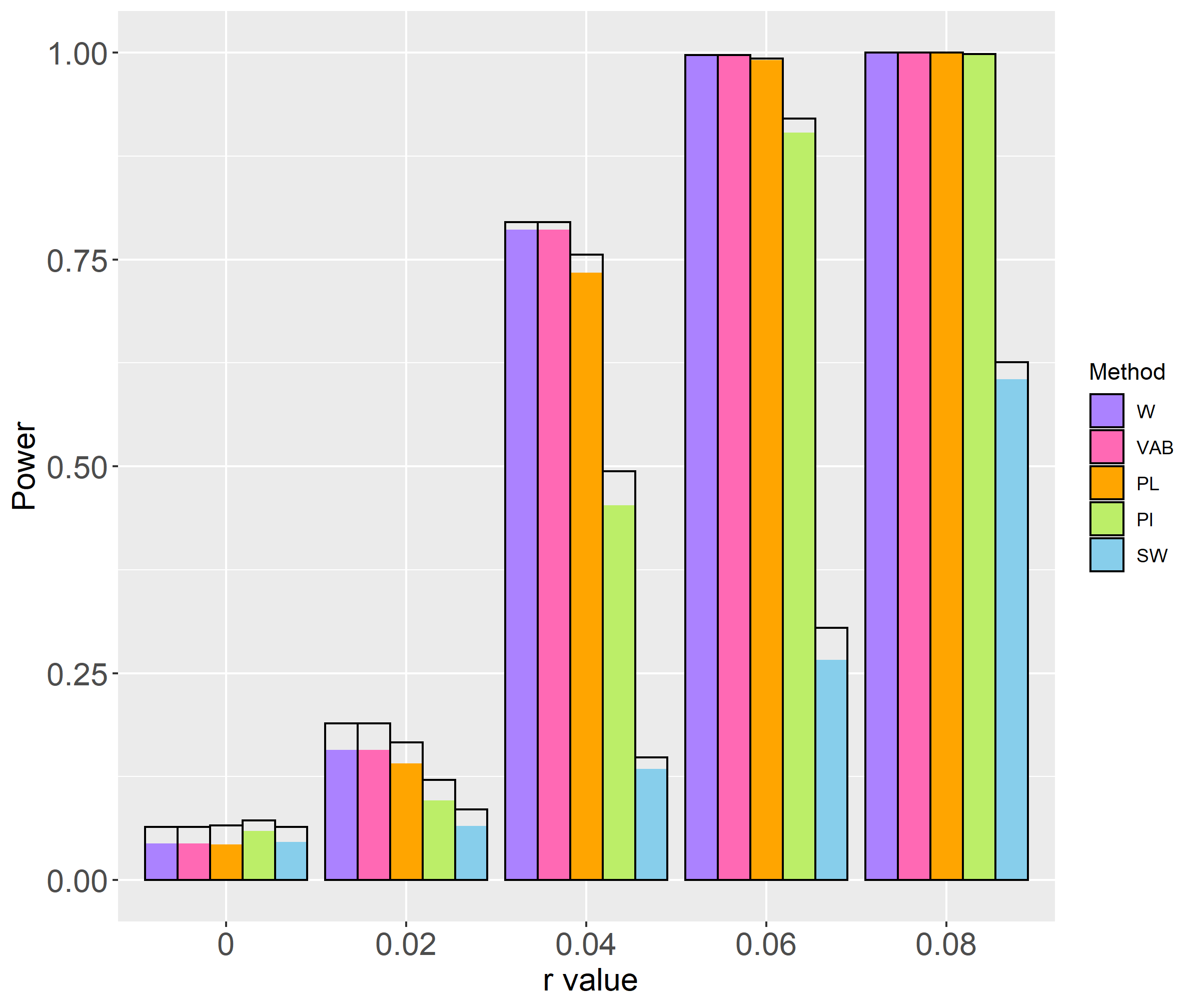}
  \vspace{-0.5cm}
  \caption{\small No noise}
  \label{fig:circle_vs_ellipse_no_noise}
\end{subfigure}\quad \quad
\begin{subfigure}[b]{.4\textwidth}
  \centering 
  \includegraphics[width=\textwidth]{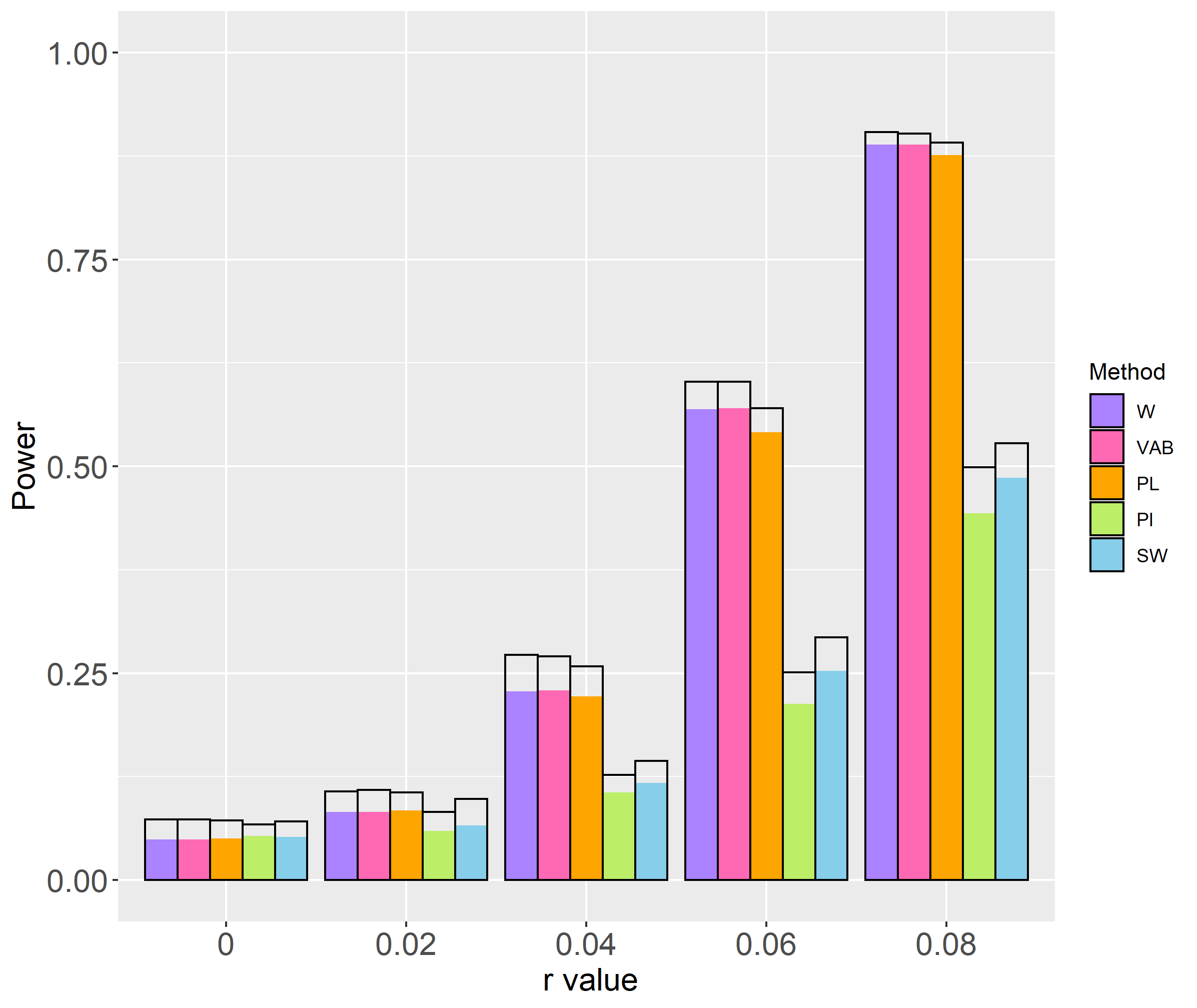}
  \vspace{-0.5cm}
  \caption{\small With noise}
  \label{fig:circle_vs_ellipse_with_noise}
\end{subfigure}
\vspace{-0.2cm}
\caption{\small Estimated powers of the permutation test (see Algorithm \ref{alg:p-value}) using the five methods (W,VAB,PL,PI,SW) for the experiment of Section \ref{subsec:circle_vs_ellipse}. PDs are computed from 2D point clouds via Vietoris-Rips filtration. Results are for homological dimension 1. The extra bars above the colored ones display the increase in power under the strong mixing approach for permuting group labels.}
\label{fig:Circle_H1.}
\end{figure}

The simulation results for homological dimension 1 are given in Figure \ref{fig:circle_vs_ellipse_no_noise} (no noise) and Figure \ref{fig:circle_vs_ellipse_with_noise} (with noise). As expected, the test powers generally increase with $r$ (or $i$). The methods W (Wasserstein) and VAB (vector of averaged Bettis) show very similar high performance, whereas the lowest powers are given by the SW (sliced Wasserstein) method. Moreover, we observe that the strong mixing procedure leads to an increase in the power of the test and it is even more pronounced when the data is noisy. More specifically, when $r=0$ (i.e. 
$H_0$ is true), under the standard permutation method, the powers (or the type-I error rates) are in the range of 4.3-5.9\% (no noise) and 4.9-5.3\% (with noise) and hence generally stay below the significance level of 5\% ($\alpha=0.05$). Under the strong mixing procedure, the type-I error rates further increase by 1.3-2.4\%. However, for $r>0$ (i.e. $H_1$ is true), the changes in the power of the test between the two permutation methods reach up to 3.9\% (no noise) and 5.6\% (with noise). The highest powers for homological dimension 0 are 11.5\% (no noise) and 14.4\% (with noise) achieved by the VAB method at $r=0.08$ (or $i=5$).

\subsubsection{Distinguishing sphere and ellipsoid}
\label{subsec:sphere_vs_ellipsoid}

The experiment of this section involves 3D point clouds and closely mimics the preceding one in Section \ref{subsec:circle_vs_ellipse}. PDs of group 1 are computed from point clouds of size 100 uniformly sampled from a unit sphere via  the alpha filtration. Point clouds for group 2 consist of 100 points sampled from the ellipsoid $x^2+y^2+z^2/(1-r)^2=1$ with $r=0.01(i-1)$, $i=1,\ldots,5$ (the index of simulation) which amounts to compressing the unit sphere by a factor of $1-r$ along the $z$-axis. Finally, the simulation is repeated by adding Gaussian noise with standard deviation $\sigma=0.05$ to all three coordinates of the points in a point cloud during the data generation process. 

\begin{figure}[t]
\centering
\begin{subfigure}[b]{.4\textwidth}
  \centering 
  \includegraphics[width=\linewidth]{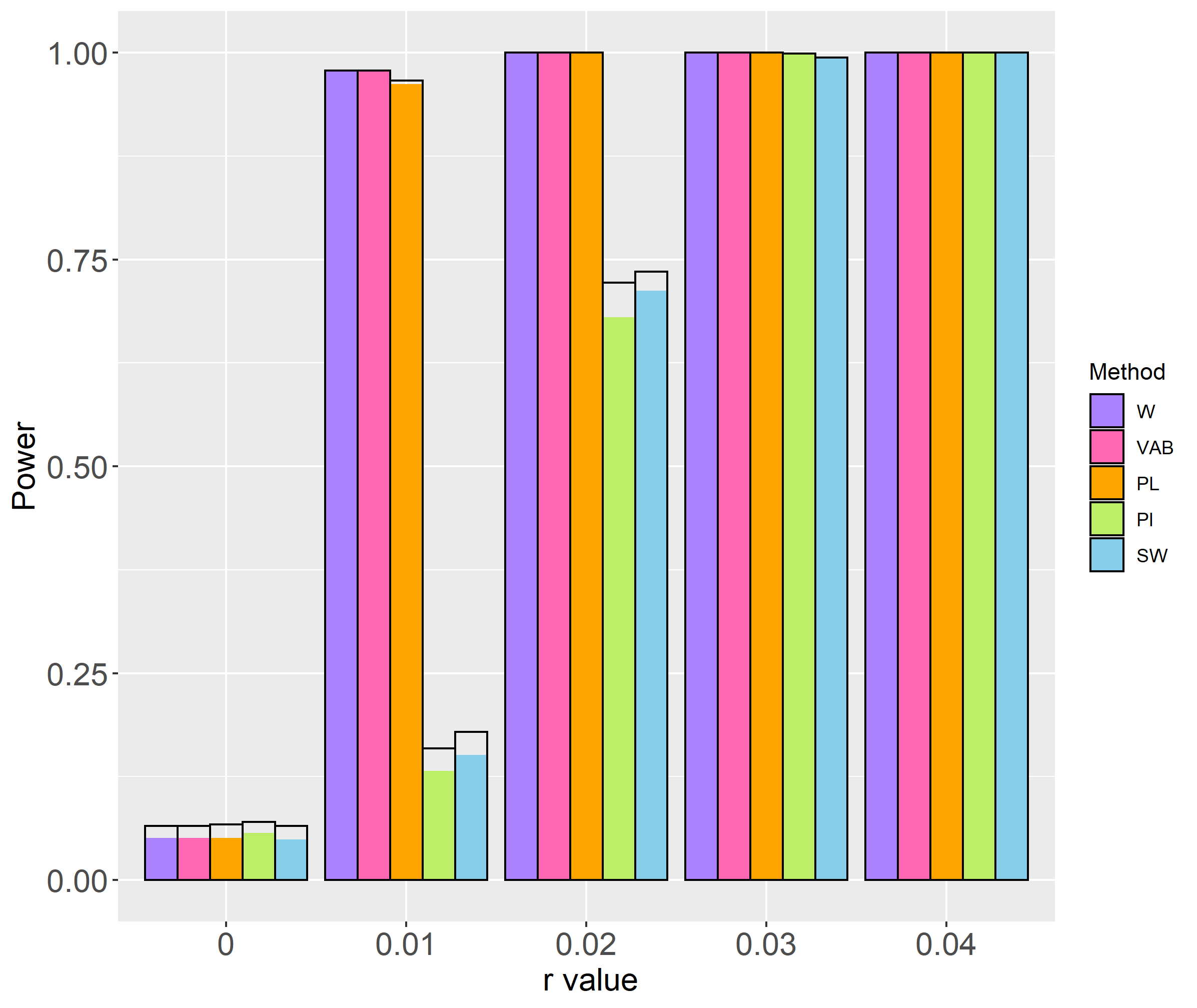}
  \vspace{-0.5cm}
  \caption{\small No noise}
  \label{fig:sphere_vs_ellipsoid_no_noise}
\end{subfigure}\quad \quad
\begin{subfigure}[b]{.4\textwidth}
  \centering 
  \includegraphics[width=\linewidth]{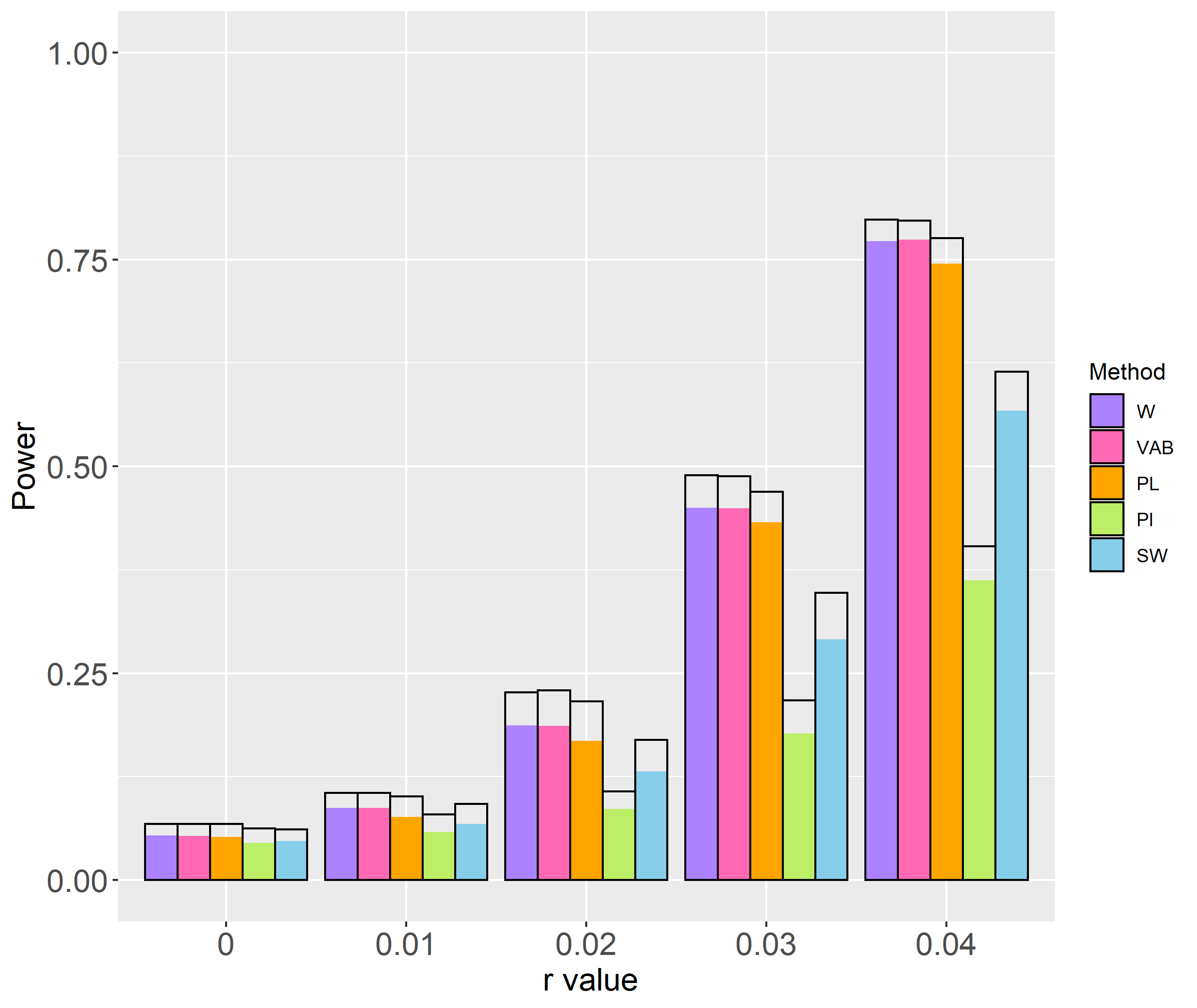}
  \vspace{-0.5cm}
  \caption{\small With noise}
  \label{fig:sphere_vs_ellipsoid_with_noise}
\end{subfigure}
\vspace{-0.2cm}
\caption{\small Estimated powers of the permutation test (see Algorithm \ref{alg:p-value}) using the five methods (W,VAB,PL,PI,SW) for the experiment of Section \ref{subsec:sphere_vs_ellipsoid}. PDs are computed from 3D point clouds via alpha complex filtration. Results are for homological dimension 2. The extra bars above the colored ones display the increase in power under the strong mixing approach for permuting group labels.}
\label{fig:Sphere_H2.}
\end{figure}

The best results are obtained for homological dimension 2 and presented in Figure \ref{fig:sphere_vs_ellipsoid_no_noise} (no noise) and Figure \ref{fig:sphere_vs_ellipsoid_with_noise} (with noise). We observe that, in the absense of noise in the data, the power for the methods W (Wasserstein), VAB (vector of averaged Bettis) and PL (persistence landscape) reach 96.2-97.7\% as early as $r=0.01$ (or $i=2$), whereas the methods PI (persistence image) and SW (sliced Wasserstein) get to that level only at $r=0.03$ (or $i=4$). In this case, due to a rapid increase of the test power, the effect of the strong mixing method is less pronounced compared to that of Section \ref{subsec:circle_vs_ellipse}. 

With the presence of noise in the data, a more gradual increase in the test power is observed reaching at most 77.4\% (B) and 79.8\% (W) by the two permutation methods respectively (see Figure \ref{fig:sphere_vs_ellipsoid_with_noise}). Here, the strong mixing leads to a maximum of 5.6\% increase in the test power. For homological dimensions 0 and 1, the highest value of the power does not exceed 12.1\% showing very limited ability to distinguish a sphere from an ellipsoid.

\subsubsection{Distinguishing persistence diagrams generated from a process}
\label{subsec:PD_from_process}

The PDs in this experiment are generated from a process (similar to one used in Section \ref{sec:cost}) rather than being computed from a point cloud through some filtration. For each simulation $i\in\{1,\ldots,5\}$, a persistence diagram $D=\{(b_j,d_j)\}_{j=1}^{50}$ of group 1 is generated from the process where $b_j\sim \hbox{Beta}(1,1)$ (or equivalently $\hbox{Unif}(0,1)$) and $d_j=b_j+z_j$ with $z_j \sim \hbox{Beta}(1,1)$. For a diagram $\tilde D=\{(\tilde b_j,\tilde d_j)\}_{j=1}^{50}$ of group 2, $b_j\sim \hbox{Beta}(1,1)$ and $d_j=b_j+z_j$ with $z_j \sim \hbox{Beta}(1,r)$, where $r=1+0.2(i-1)$, $i=1,\ldots,5$. As in the previous two experimental settings, when $r=0$ (or $i=1$) the two processes behind the generation of PDs are the same (i.e. the $H_0$ hypothesis is true) and as $r$ increases the processes begin to diverge from one another. For the noisy case, Gaussian noise with standard deviation $\sigma=0.1$ is added to both the birth $b$ and death $d$ coordinates of each point in a PD. If, as a result, $b>d$, then the point $(b,d)$ is replaced by $(b,b)$ to ensure that no point lies below the diagonal $\Delta=\{(b,d)|b=d\}$.                

\begin{figure}[t]
\centering
\begin{subfigure}[b]{.4\textwidth}
  \centering 
  \includegraphics[width=\linewidth]{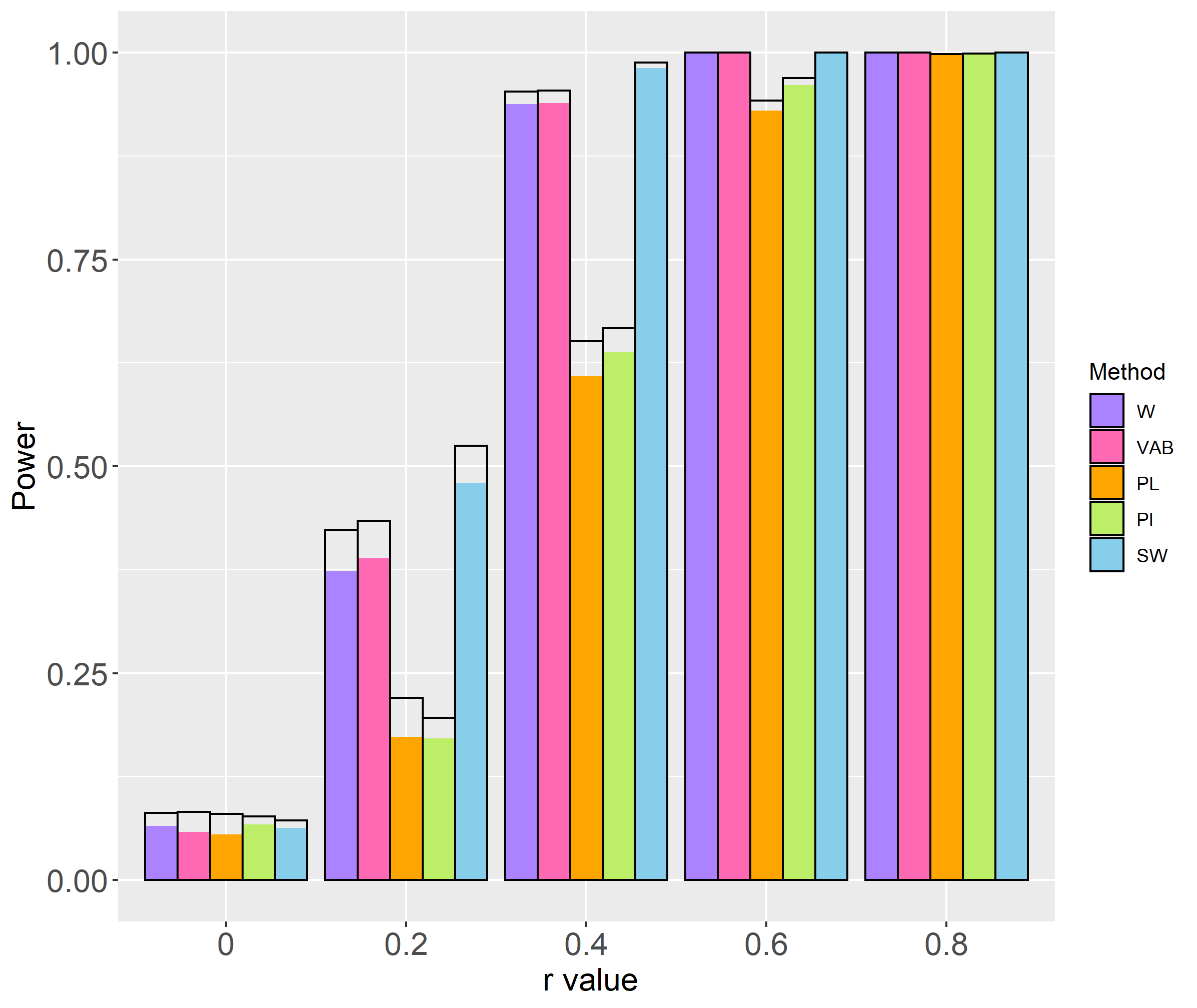}
  \vspace{-0.5cm}
  \caption{\small No noise}
  \label{fig:PD_from_process_no_noise}
\end{subfigure} \quad \quad
\begin{subfigure}[b]{.4\textwidth}
  \centering 
  \includegraphics[width=\linewidth]{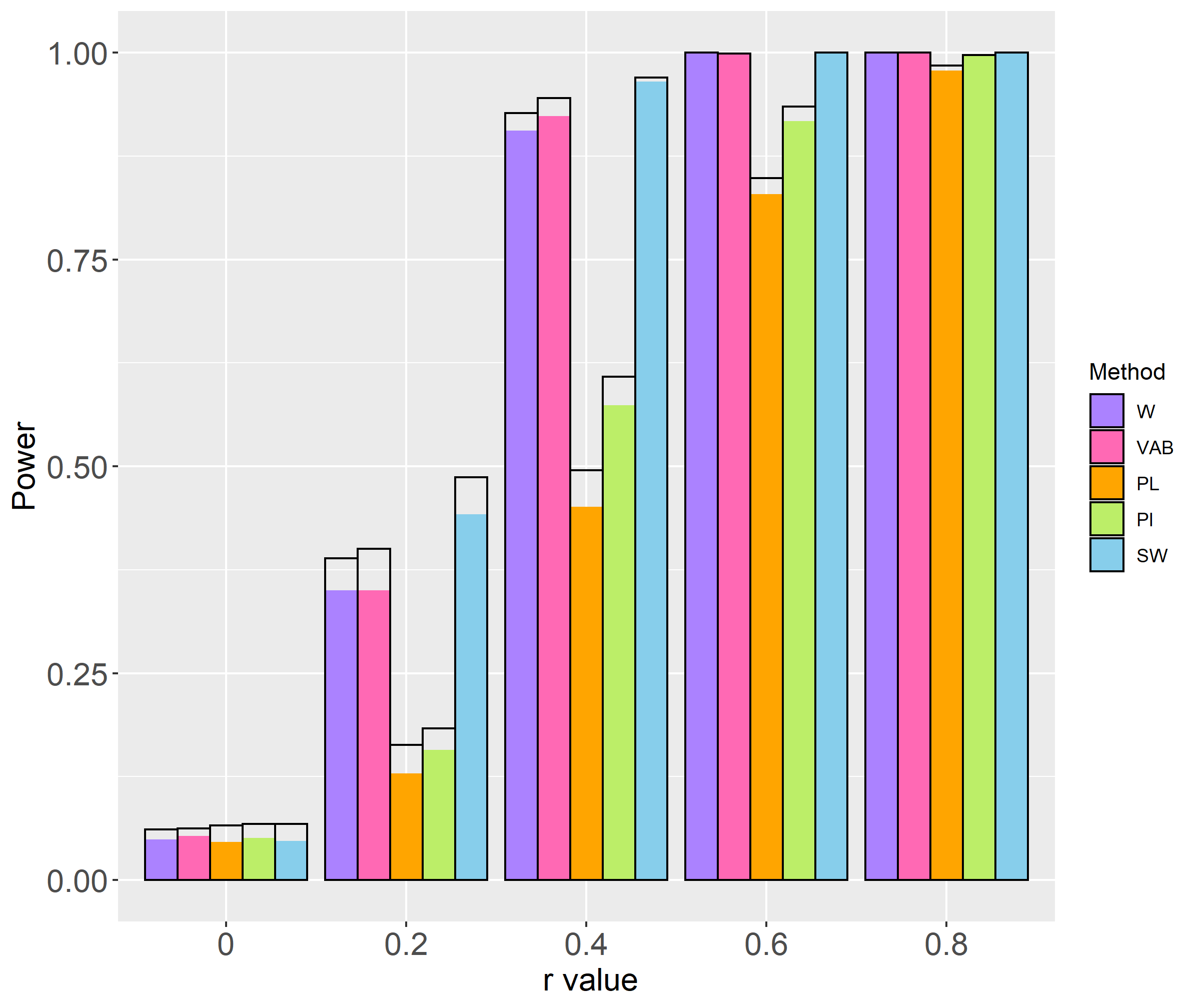}
  \vspace{-0.5cm}
  \caption{\small With noise}
  \label{fig:PD_from_process_with_noise}
\end{subfigure}
\vspace{-0.2cm}
\caption{\small Estimated powers of the permutation test (see Algorithm \ref{alg:p-value}) using the five methods (W,VAB,PL,PI,SW) for the experiment of Section \ref{subsec:PD_from_process}. PDs are generated from a process. The extra bars above the colored ones display the increase in power under the strong mixing approach for permuting group labels.}
\label{fig:Beta_H0.}
\end{figure}

The results are given Figures \ref{fig:PD_from_process_no_noise} and \ref{fig:PD_from_process_with_noise}. Unlike in the previous two experiments, the method SW (sliced Wasserstein) provides the highest test powers. When the two processes are still relatively hard to distinguish (e.g. $r=0.2$ and $r=0.4$), the effect of the strong mixing is observed the most (up to 5\%) and the methods PL (persistence landscape) and PI (persistence image) have the lowest powers compared to the other three methods. 

\subsubsection{Distinguishing simulated graphs}
\label{Distinguishing graphs from simulated data}

The final experiment in this section involves computing PDs from simulated graphs using the lower-star filtration. More specifically, first we sample 100 points from the Dirichlet distribution with parameter $\alpha\in\mathbb{R}^3$ and generate a graph whose nodes are represented by the sampled points and an edge is added between nodes $x=(x_1,x_2,x_3)$ and $y=(y_1,y_2,y_3)$ with probability $p=x_1y_1+x_2y_2+x_3y_3$ (dot product between $x$ and $y$). This graph is called a \emph{random dot product graph} and has been proven useful to simulate of social networks \cite{kraetzl2005random}. Next, viewing the graph as a one-dimensional simplicial complex, we increase its dimension to two by adding triangles (2-simplices) formed by the edges. Finally, we compute the PD of the lower-star filtration induced by $g(x)=-\sum_{i=1}^3 x_i\log(x_i)$ (cross-entropy) where $x=(x_1,x_2,x_3)$ represents a node. 

All graphs of group 1 are generated using $\alpha=(1.5,1.5,1.5)$. For each simulation $i \in \{1,\ldots,5\}$, the graphs of group 2 are produced from the Dirichlet distribution with $\alpha=(1.5,1.5,1.5+0.4(i-1))$. We repeat the experiment by adding Gaussian noise with standard deviation $\sigma=0.1$ to the values of the function $g$ (node attributes). 

\begin{figure}[t]
\centering
\begin{subfigure}[b]{.4\textwidth}
  \centering 
  \includegraphics[width=\linewidth]{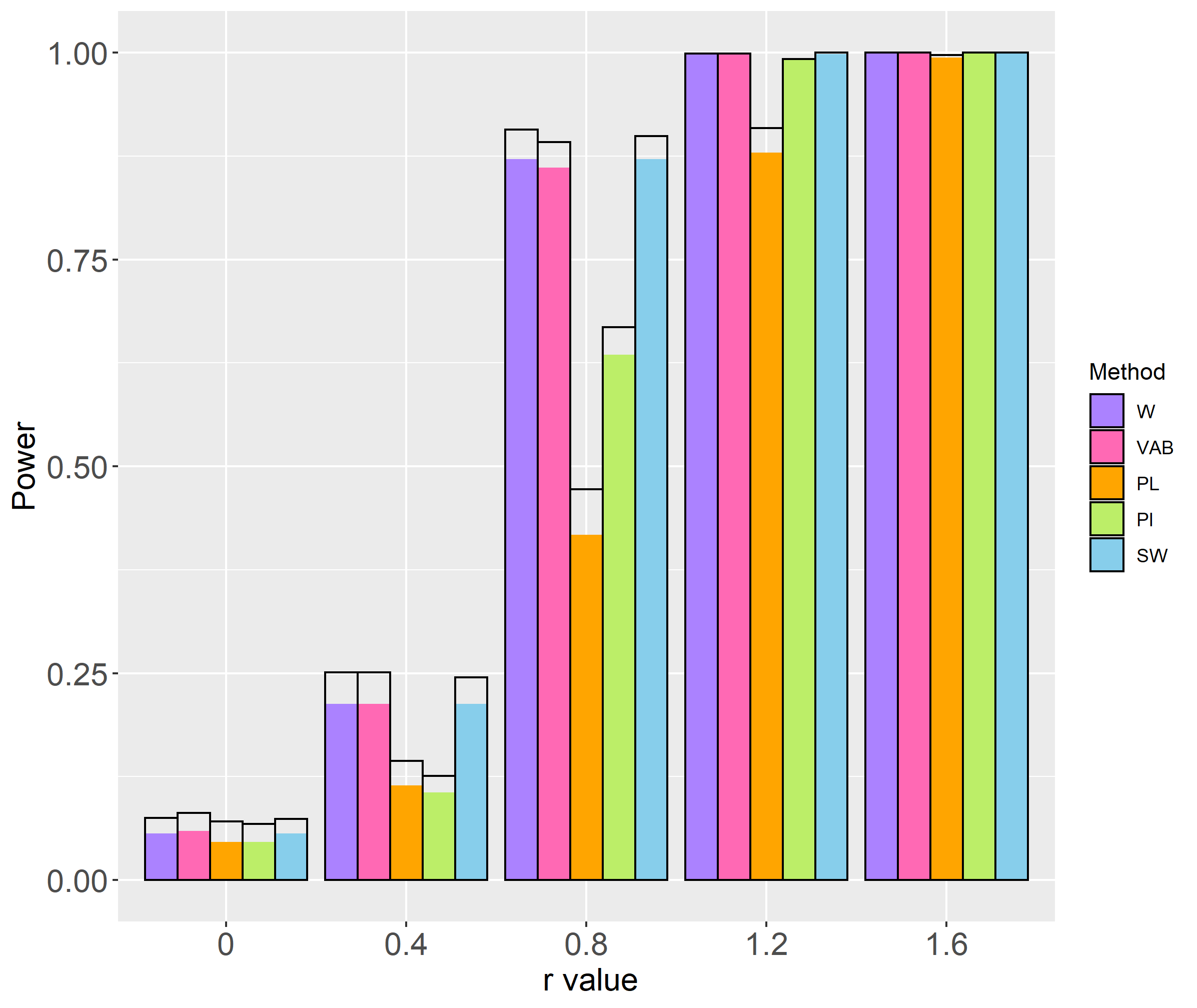}
  \vspace{-0.5cm}
  \caption{\small No noise}
  \label{fig:Graph-NoNoise-H1}
\end{subfigure}\quad \quad
\begin{subfigure}[b]{.4\textwidth}
  \centering 
  \includegraphics[width=\linewidth]{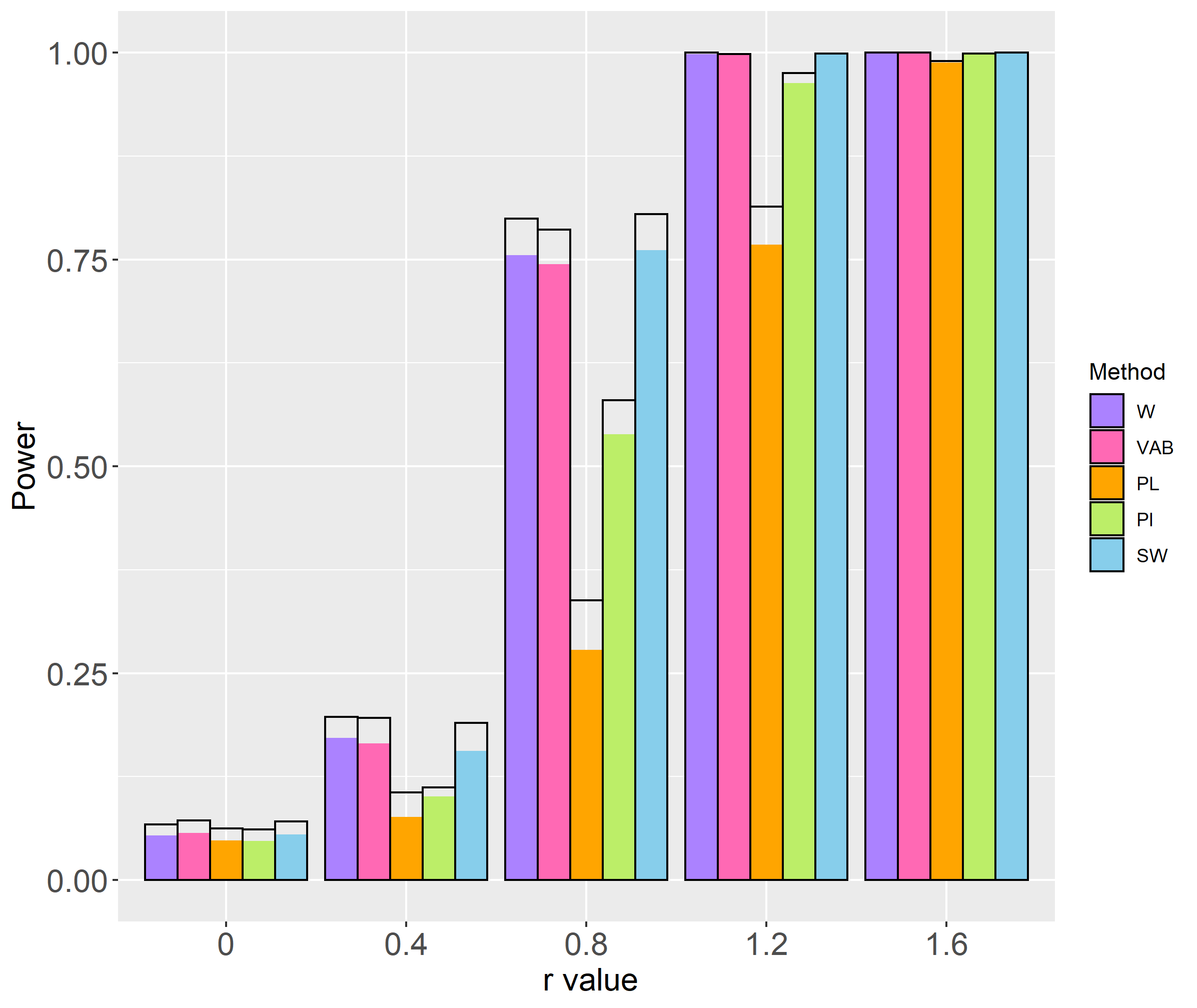}
  \vspace{-0.5cm}
  \caption{\small With noise}
  \label{fig:Graph-Noise-H1}
\end{subfigure}
\vspace{-0.2cm}
\caption{\small Estimated powers of the permutation test (see Algorithm \ref{alg:p-value}) using the five methods (W,VAB,PL,PI,SW) for the experiment of Section \ref{Distinguishing graphs from simulated data}. PDs are computed from graphs via lower-star filtration. Results are for homological dimension 1. The extra bars above the colored ones display the increase in power under the strong mixing approach for permuting group labels.}
\label{fig:Graph_H1.}
\end{figure}

The best results, observed for homological dimension 1, are given in Figures \ref{fig:Graph-NoNoise-H1} and \ref{fig:Graph-Noise-H1}. In this experiment, the methods W (Wasserstein) and VAB (vector of averaged Betttis) show the highest test powers for all five simulations. Again, the effect of the strong mixing technique is most pronounced for $r=0.4$ and $r=0.8$ (when the two groups are still relatively difficult to tell apart).

\subsection{Hypothesis testing on real data}
\label{Performance comparison using hypothesis testing with real data}

In the final section, we conduct two experiments on hypothesis testing which involve PDs computed from real data. In the first experiment, we consider point clouds sampled for the surfaces of ten 3D non-rigid toys and compute PDs using the alpha filtration. The second experiment deals with PDs computed from real graph data using the lower-star filtration induced by various functions for node attributes. 

\subsubsection{Distinguishing 3D non-rigid toys}
\label{sec:shrec}
\begin{figure}[t]
\centering
\begin{subfigure}[b]{.37\textwidth}
  \centering 
  \includegraphics[width=0.8\linewidth]{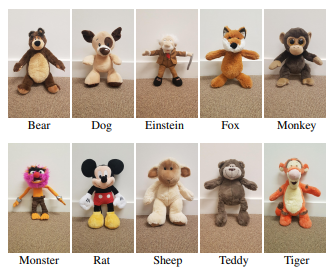}
  \caption{\small The ten non-rigid toys used to generate 3D point clouds}
  \label{fig:Shrec_toys}
\end{subfigure}\quad \quad
\begin{subfigure}[b]{.37\textwidth}
  \centering 
  \includegraphics[width=1.1\linewidth]{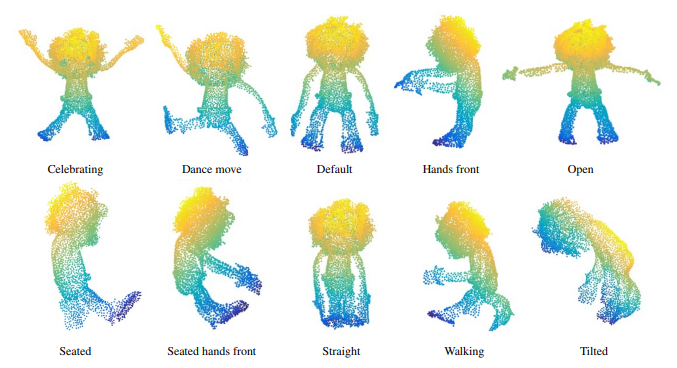}
  \caption{\small 3D point clouds of the Monster toy in various poses}
  \label{fig:Shrec_poses}
\end{subfigure}
\vspace{-0.2cm}
\caption{\small Toys used for the experiment of Section \ref{sec:shrec} \cite{limberger2017shrec}.} 
\label{fig:Shrec data}
\end{figure}

The dataset for this experiment contains
point clouds obtained from ten non-rigid toys \cite{limberger2017shrec}. Each toy is scanned in ten different poses resulting in a total of 100 point clouds (of size 4000 on average) sampled uniformly from the toy surface scans (see Figures \ref{fig:Shrec_toys} and \ref{fig:Shrec_poses}). To compute PDs, we resort to the alpha filtration, rather than the Vietoris-Rips filtration, as it has much lower run-time cost for this dataset. The best results are obtained for homological dimension 2 which tracks the evolution of voids (or holes of dimension two) along the filtration. As a note, in this experiment the PDs are much larger in size (on average 4000, 7500 and 1200 for homological dimensions 0, 1 and 2 respectively) compared to those of Section \ref{sec:HT_simulated_data} which leads to significantly higher run-time cost of computing Wasserstein distances for Algorithm \ref{alg:p-value}.


First, we compare points clouds sampled from the same toy. Thus, for each toy there are ten point clouds which we randomly split into two groups of equal size. Algorithm \ref{alg:p-value} is run with $N=\binom{10}{5}=252$ permutations and the standard mixing procedure for permuting the group labels. Table \ref{table:Shrec-similar} contains the p-values obtained for each toy and each of the five methods (W, VAB, PI, PL, SW) for computing distance matrices. We see from Table \ref{table:Shrec-similar} that since the two groups of PDs come from the same toy, the null hypothesis $H_0$ is rejected only 8\% of the time (4 out 50 comparisons) which is close to the used significance level of 5\%.

\begin{table}[t]
 	\centering
	\begin{tabular}{cccccc}
		\hline
		 & W  & VAB  & PL  & PI  & SW\\ 
		\hline
		Toy 1 & 0.241 &	0.601 &	0.502 &	0.233 &	0.320\\
		Toy 2 & 0.458 &	0.601 &	0.138 &	0.170 &	0.352 \\
		Toy 3 & 0.344 &	0.545 &	0.798 &	0.265 &	0.498\\
		Toy 4 & 0.202 &	0.625 &	0.352 &	0.375 &	0.253\\
		Toy 5 & 0.138 &	0.352 &	\textbf{0.036} & 0.075 &	0.130\\
		Toy 6 & 0.119 &	0.083 &	\textbf{0.043} &	0.281 &	0.067 \\
		Toy 7 & 0.530 &	0.265 &	0.099 &	0.067 &	0.628 \\
		Toy 8 & 0.249 &	0.557 &	0.379 &	0.249 &	0.253\\
		Toy 9 & \textbf{0.036} &	0.715 &	0.051 &	0.099 &	\textbf{0.043}\\
		Toy 10 & 0.577 & 0.059 & 0.482 &	0.565 & 0.296\\
		\hline
		\end{tabular}
	\vspace{0.1cm}
	\caption{\small Estimated p-values of the permutation test (see Algorithm \ref{alg:p-value}) using the five methods (W,VAB,PL,PI,SW) for the experiment of Section \ref{sec:shrec}, where point clouds are sampled from the same toy.}
	\label{table:Shrec-similar}
 \end{table} 

Next, we compare point clouds obtained from different toys. Algorithm \ref{alg:p-value} is applied to all possible $\binom{10}{2}=45$ comparisons (with $N=10000$ permutations) using the five methods for computing distance matrices. Out of the $45\times 5 = 225$ computed p-values, all except one turn out be below $\alpha=0.05$ (we fail to reject the $H_0$ hypothesis only once for the PL method). Thus, in all but one case, the hypothesis testing procedure correctly distinguishes the pairs toys compared. 

\subsubsection{Distinguishing real graphs}
\label{Experiments with real graph data}
The experiments of this section utilize datasets from TUDataset - a collection of benchmark datasets for graph classification and regression \cite{KKMMN2016}. The TUDataset collection contains more than 120 datasets from various application domains. We use 13 of those datasets where the graphs are classified into two groups (see Table \ref{table:data-realgraph} for their names and basic graph summaries). Table \ref{table:data-realgraph} shows that the graphs vary in terms of the number of nodes, group sizes and the balance between them.

\begin{table}[t]
\centering
      \begin{tabular}{lcccc}
		\hline
		Dataset &	Group 1 size &	Group 2 size & Avg \# of nodes &	Avg \# of edges\\ 
		\hline
		AIDS &	400	 & 1600 &	15.69 &	16.20 \\
		BZR &	319 &	86 &	35.75 &	38.36 \\
        COX2 &	365 &	102 &	41.22 &	43.45 \\
        DHFR &	295 &	461 &	42.43 &	44.54 \\
        DD &	691 &	487 &	284.3 &	715.7 \\
        IMDB-BINARY &	500 &	500 &	19.77 &	96.53 \\
        KKI &	37 &	46 &	26.96 &	48.42 \\
        MUTAG &	63 &	125 &	17.93 &	19.79 \\
        OHSU &	35 &	44 &	82.01 &	199.7 \\
        Peking\_1 &	49 &	36 & 39.31 &	77.35 \\
        PTC\_FM &	206 &	143 &	14.11 &	14.48 \\
        PROTEINS &	663 &	450 &	39.06 &	72.82 \\
        REDDIT-BINARY &	1000 &	1000 &	429.6 &	497.8  \\
       \hline
	\end{tabular}
	\vspace{0.1cm}
	\caption{\small Graph datasets used for the experiment of Section \ref{Experiments  with real graph data}.}
	\label{table:data-realgraph}
 \end{table} 

We compute PDs of homological dimensions 0 and 1 using the lower-start filtration induced by the node attributes of degree, betweenness and closeness. In case the graphs are further supplied with (external) node attributes, we utilize those attributes as well to produce the filtration. We apply Algorithm \ref{alg:p-value} with $N=10000$ permutations and both standard and strong mixing procedures for permuting the class labels. Due to a high relative run-time cost of computing the Wasserstein distances, we calculate them only if the total number of graphs in both groups does not exceed 100 (which amounts to three datasets). Table \ref{table:pvals-realgraph} provides the hypothesis testing results for each dataset and method used to compute the distance matrix. The results show that the hypothesis testing procedure distinguishes the two groups of graphs in the majority of cases based on at least one of the node attribute functions used. As in Section \ref{Distinguishing graphs from simulated data}, better results are obtained mostly for homological dimension 1. All methods fail to tell apart the graphs in the BZR and KKI datasets at the significance level of 5\%. The methods W (Wasserstein) (if available), VAB (vector of averaged Bettis) and SW (sliced Wasserstein) provide almost identical results. The strong mixing procedure helps improve the results but only in few cases. Overall, all the five methods demonstrate comparable performance. 

\begin{table}[t]
\centering
      \begin{tabular}{lccccc}
		\hline
		Dataset (hom dim used) &	W  & VAB & PL &	PI & SW\\ 
		\hline
		AIDS (0) & - &     $\times$,$\times$,$\times$,\checkmark &
        $\times$,$\times$,$\times$,$\times$ &
        \checkmark,\checkmark,\checkmark,\checkmark &
        $\times$,$\times$,$\times$,$\times$ \\
		BZR (0,1)	&	- &     $\times$,$\times$,$\times$,$\times$ &
        $\times$,$\times$,$\times$,$\times$ &
        $\times$,$\times$,$\times$,$\times$ &
        $\times$,$\times$,$\times$,$\times$  \\
        COX2 (1)& -&     \checkmark,\checkmark,\checkmark,\checkmark&
        $\times$,\checkmark,\checkmark,$\times$ &
        \checkmark,\checkmark,\checkmark,$\times$ &
        \checkmark,\checkmark,\checkmark,\checkmark \\
        DHFR (1)&-	&
        \checkmark,\checkmark,\checkmark,\checkmark&
        \checkmark,\checkmark,\checkmark,$\times$ &
        \checkmark,\checkmark,$\times$,$\times$ &
        \checkmark,\checkmark,\checkmark,\checkmark \\
        DD (1) & - &
        \checkmark,\checkmark,\checkmark,\hbox{ - } &
        \checkmark,\hbox{ $\star$ },\checkmark,\hbox{ - } &
        \checkmark,\checkmark,\checkmark,\hbox{ - } &
        \checkmark,\checkmark,\checkmark,\hbox{ - } \\
        IMDB-BINARY (1)	& - &
        \checkmark,\checkmark,\checkmark,\hbox{ - } &
        \checkmark,\checkmark,\checkmark,\hbox{ - } &
        \checkmark,\checkmark,\checkmark,\hbox{ - } &
        \checkmark,\checkmark,\checkmark,\hbox{ - } \\
        KKI (1)	&
        $\times$,$\times$,$\times$,\hbox{ - }&     $\times$,$\times$,$\times$,\hbox{ - }&
        \hbox{$\star$ },$\times$,$\times$,\hbox{ - }&     $\times$,$\times$,$\times$,\hbox{ - }&
        $\times$,$\times$,$\times$,\hbox{ - } \\
        MUTAG (1) & - &
        \checkmark,\checkmark,\checkmark,\hbox{ - } &
        $\times$,\checkmark,\checkmark,\hbox{ - } &
        \checkmark,\checkmark,\checkmark,\hbox{ - } &
        \checkmark,\checkmark,\checkmark,\hbox{ - }  \\
        OHSU (0) &
        $\times$,\hbox{ $\star$ },$\times$,\hbox{ - }&    $\times$,\hbox{ $\star$ },$\times$,\hbox{ - }&
        $\times$,$\times$,$\times$,\hbox{ - }&
        $\times$,$\times$,$\times$,\hbox{ - }&
        $\times$,\hbox{ $\star$ },$\times$,\hbox{ - } \\
        Peking\_1 (0) &
        \checkmark,\hbox{ $\star$ },$\times$,\hbox{ - }&       \checkmark,\hbox{ $\star$ },$\times$,\hbox{ - }&
        $\times$,$\times$,$\times$,\hbox{ - }&
        $\times$,$\times$,$\times$,\hbox{ - }&
        \checkmark,\hbox{ $\star$ },$\times$,\hbox{ - } \\
        PTC\_FM (1) &-&     \checkmark,$\times$,$\times$,\hbox{ - }&
        \checkmark,\checkmark,\hbox{ $\star$ },\hbox{ - } &
        \checkmark,$\times$,$\times$,\hbox{ - }&
        \checkmark,$\times$,$\times$,\hbox{ - } \\
        PROTEINS (1) &-&     \checkmark,\checkmark,\checkmark,\checkmark&
        \checkmark,\checkmark,\checkmark,\checkmark &
        \checkmark,\checkmark,\checkmark,\checkmark &
        \checkmark,\checkmark,\checkmark,\checkmark \\
        REDDIT-BINARY (1) &-&     \checkmark,\checkmark,\checkmark,\hbox{ - } &
        \checkmark,\checkmark,\checkmark,\hbox{ - }  &
        \checkmark,\checkmark,\checkmark,\hbox{ - }  &
        \checkmark,\checkmark,\checkmark,\hbox{ - }  \\
       \hline
	\end{tabular}
	\vspace{0.1cm}
	\caption{\small Hypothesis testing results for real graph data using the standard permutation procedure. For each method, we report four results indicated by the symbols \checkmark, \hbox{ $\star$ } and $\times$ (depending on whether the p-value is less than 0.05, between 0.05 and 0.1 or greater than 0.1) where the node attributes are respectively based on the \emph{degree, betweenness, closeness or externally provided}. For example, \checkmark,\checkmark,$\times$,$\times$ means that the p-value is less than 0.05 for the degree and betweenness and greater than 0.1 for the closeness and the case when node attributes are externally provided. The symbol \hbox{ "-" } means the result is not available. }
	\label{table:pvals-realgraph}
 \end{table}

\section{Conclusion and directions for future research}
\label{sec:conclusion}
The paper by \cite{robinson2017hypothesis} is one of the early and important works on hypothesis testing for TDA. This approach is based on the permutation test where persistence diagrams are regarded as data observations and the associated loss function (\ref{Fpq}) is defined using the sum of within-class pairwise Wasserstein distances between the diagrams. However, in the settings where persistence diagrams are large in size and in number, computing the Wasserstein distances may have high computational cost limiting the application of the permutation test. The present work proposes to replace the Wasserstein distances between persistence diagrams with the $L_p$ distances between vector summaries extracted from the diagrams. In this context, though we specifically focus on exploring the utility of the vectorized Betti function as an alternative summary to working with persistence diagrams directly, our approach is applicable to any other vector summary. 

The Betti function, or more generally any univariate summary function extracted from a persistence diagram, is typically vectorized by evaluating it at each point of a super-imposed grid of scale values. We propose a different vectorization scheme for the Betti function by averaging it between two consecutive scale values via integration. The new vector summary is called a \emph{vector of averaged Bettis} (VAB) and can be used in situations where informative low-dimensional topological vector summaries are desired. Moreover, we prove stability results for the Betti function and its vectorized form VAB with respect to the 1-Wasserstein distance.   

In our experiments, we apply the hypothesis testing procedure of \cite{robinson2017hypothesis} to test whether the process or underlying shape behind two groups of persistence diagrams is the same (per homological dimension) by employing five methods -- the 1-Wasserstein distance (baseline), sliced Wasserstein distance and the $L_1$ distance between the vector summaries of VAB (our focus), persistence landscape (PL) and persistence image (PI) -- to compute the loss function. We use both synthetic and real datasets of various types and we find that the results obtained based on the Wasserstein distance and VAB are very similar and outperform the other three methods in the majority of cases. In almost all cases, persistence diagrams corresponding to the highest available homological dimension lead to the best results.

In practice, to avoid a high computational cost, Algorithm \ref{alg:p-value} implementing the permutation test in \cite{robinson2017hypothesis} often uses only a fraction of permutations uniformly sampled from all possible permutations to estimate the p-value. In our experiments, in addition to the standard permutation procedure, we introduce and explore a new shuffling procedure which considers only those permutations which mix the labels of the two groups the most. The experimental results show that the proposed permutation procedure leads to increased power of the test. This increase is more pronounced when the parameter quantifying the difference between the two groups of persistence diagrams is relatively small (i.e. when the two groups of diagrams are harder to distinguish). However, the observed improvement of the test power comes at the cost of minor inflation of the type-I error. 

The loss function (\ref{alg:p-value}) assigns equal weights to both groups of persistence diagrams regardless of their sizes. After obtaining our main results, we investigated the performance of the loss function when the weights are adjusted according to the group sizes. The preliminary experiments with the modified loss function on the TUDataset of real graphs (see Section \ref{Experiments with real graph data}) result in higher percentage of correct decisions. For example, the graphs in the AIDS and BZR datasets (both of which are imbalanced) are easily distinguished by the methods VAB, PL and SW for almost all node attribute functions and homological dimensions used. We plan to explore theoretical properties of the new permutation procedure and the modified loss function in our future research. 

\bibliographystyle{unsrt}  
\bibliography{literature.bib}

\end{document}